%% file: Main.tex
\documentclass[lettersize,journal]{IEEEtran}
\usepackage{amsmath,amsfonts}
\usepackage{algorithmic}
\usepackage{algorithm}
\usepackage{array}
\usepackage[caption=false,font=normalsize,labelfont=sf,textfont=sf]{subfig}
\usepackage{textcomp}
\usepackage{stfloats}
\usepackage{url}
\usepackage{verbatim}
\usepackage{graphicx}
\usepackage{cite}
\newcommand{\minisection}[1]{\vspace{0.02in} \noindent {\bf #1}\ }
\usepackage[svgnames]{xcolor}
\usepackage{subcaption}

\usepackage{algorithmic}
\usepackage{algorithm}
\usepackage{hyperref}
\usepackage{cleveref}
\usepackage{booktabs}

\begin{document}

\let\WriteBookmarks\relax
\def\floatpagepagefraction{1}
\def\textpagefraction{.001}

\title{AViTMP: A Tracking-Specific Transformer for Single-Branch Visual Tracking}

\author{Chuanming Tang, Kai Wang, Joost van de Weijer, Jianlin Zhang, Yongmei Huang
\thanks{Chuanming Tang is with University of Chinese Academy of Sciences, Beijing, 108408, China; also with Key Laboratory of Optical Engineering, Institute of Optics and Electronics, Chinese Academy of Sciences, Chengdu, 610209, China; also with Computer Vision Center, Barcelona, 08193, Spain (e-mail: tangchuanming19@mails.ucas.ac.cn)}
\thanks{Kai Wang and Joost van de Weijer are with Computer Vision Center, Universitat Autonoma de Barcelona, Barcelona, 08193, Spain (e-mail: \{kwang; joost\}@cvc.uab.es)}
\thanks{Jianlin Zhang and Yongmei Huang are with Key Laboratory of Optical Engineering, Institute of Optics and Electronics, Chinese Academy of Sciences, Chengdu, 610209, China  (e-mail: \{jlin; huangym\}@ioe.ac.cn)}

\thanks{Corresponding authors: Kai Wang, Yongmei Huang}
}

\markboth{IEEE Transactions on Intelligence Vehicles}%
{Shell \MakeLowercase{\textit{et al.}}: A Sample Article Using IEEEtran.cls for IEEE Journals}


\maketitle

\begin{abstract}
Visual object tracking is a fundamental component of transportation systems, especially for intelligent driving. Despite achieving state-of-the-art performance in visual tracking, recent single-branch trackers tend to overlook the weak prior assumptions associated with the Vision Transformer (ViT) encoder and inference pipeline in visual tracking. Moreover, the effectiveness of discriminative trackers remains constrained due to the adoption of the dual-branch pipeline. To tackle the inferior effectiveness of vanilla ViT, we propose an Adaptive ViT Model Prediction tracker (AViTMP) to design a customised tracking method. This method bridges the single-branch network with discriminative models for the first time. Specifically, in the proposed encoder AViT encoder, we introduce a tracking-tailored Adaptor module for vanilla ViT and a joint target state embedding to enrich the target-prior embedding paradigm. Then, we combine the AViT encoder with a discriminative transformer-specific model predictor to predict the accurate location. Furthermore, to mitigate the limitations of conventional inference practice, we present a novel inference pipeline called CycleTrack, which bolsters the tracking robustness in the presence of distractors via bidirectional cycle tracking verification. In the experiments, we evaluated AViTMP on eight tracking benchmarks for a comprehensive assessment, including LaSOT, LaSOTExtSub, AVisT, etc. The experimental results unequivocally establish that, under fair comparison, AViTMP achieves state-of-the-art performance, especially in terms of long-term tracking and robustness. The source code will be released at \url{https://github.com/Tchuanm/AViTMP}.

\end{abstract}

\begin{IEEEkeywords}
Visual Tracking, Single-branch Tracker, Discriminative Tracker, Cycle Consistency
\end{IEEEkeywords}

\section{Introduction}
\IEEEPARstart{G}{eneric} object visual tracking is a significant challenge in computer vision, especially for current transportation systems. It can provide location and direction information for the target of interest, which is particularly beneficial for autonomous vehicles.
Unlike multi-object tracking~\cite{gunduz2019efficient}, single-object tracking is a category-ignorant task that can track any kind of object based on a prior bounding box.
It involves the estimation of the target's position in a search frame based on the target bounding box of the initial frame (also called the template frame).
\textcolor{black}{
In the field of intelligent vehicles, visual tracking poses a substantial challenge, particularly in transportation scenarios. Despite notable progress in tracking technology, numerous challenges persist, including vehicle occlusions, camera motion blur, and the potential confusion arising from similar vehicles and pedestrians. These challenges have implications for the effectiveness of visual tracking algorithms in various applications, such as autonomous driving~\cite{rangesh2019no,tang2023transformer}, intelligent surveillance~\cite{javed2022visual,hoffmann2020real}, and advanced traffic systems~\cite{liang2021deep}. }

Among the prevailing tracking techniques, dual-branch trackers (discriminative and Siamese trackers) and single-branch trackers stand out as two prominent pipelines. Dual-branch trackers process the template and search frame in separate branches, whereas single-branch methods process them jointly, allowing for early interaction. In discriminative approaches~\cite{dimp,kys,atom,danelljan2015learning,danelljan2016beyond}, the target model localizes the target position by minimizing a discriminative objective function. These discriminative trackers are all based on two weight-shared convolutional backbones in a dual-branch pipeline. 
However, compared to recently developed transformer backbones~\cite{visiontransformer2020,wu2021cvt}, CNN networks are found to be inferior in feature extraction and context-information modeling.
Moreover, the fusion of the two branches can lead to further information loss.

Siamese trackers~\cite{siamfc,chen2021transt,siamrpn,stark} learn the similarity matrix to classify and locate the foreground and background based on two branches of feature matching.
Currently, there has been a surge in the popularity of single-branch transformer-based trackers~\cite{chen2022SimTrack,chen2023seqtrack,mixformer,xie2022sbt,ye2022OSTrack}. Different from dual-branch (\textit{external cross-matching}), these frameworks apply a straightforward single-branch Vision Transformer (ViT~\cite{visiontransformer2020}) to perform \textit{internal cross-matching}. 
Typical single-branch trackers~\cite{chen2022SimTrack,chen2023seqtrack,ye2022OSTrack} follow a similar pipeline where search and template frames are concatenated together. They then directly employ vanilla ViT to facilitate information interaction by means of self-attention blocks across multiple frames.

\begin{figure}[t]
\centering 
\includegraphics[width=0.99\columnwidth]{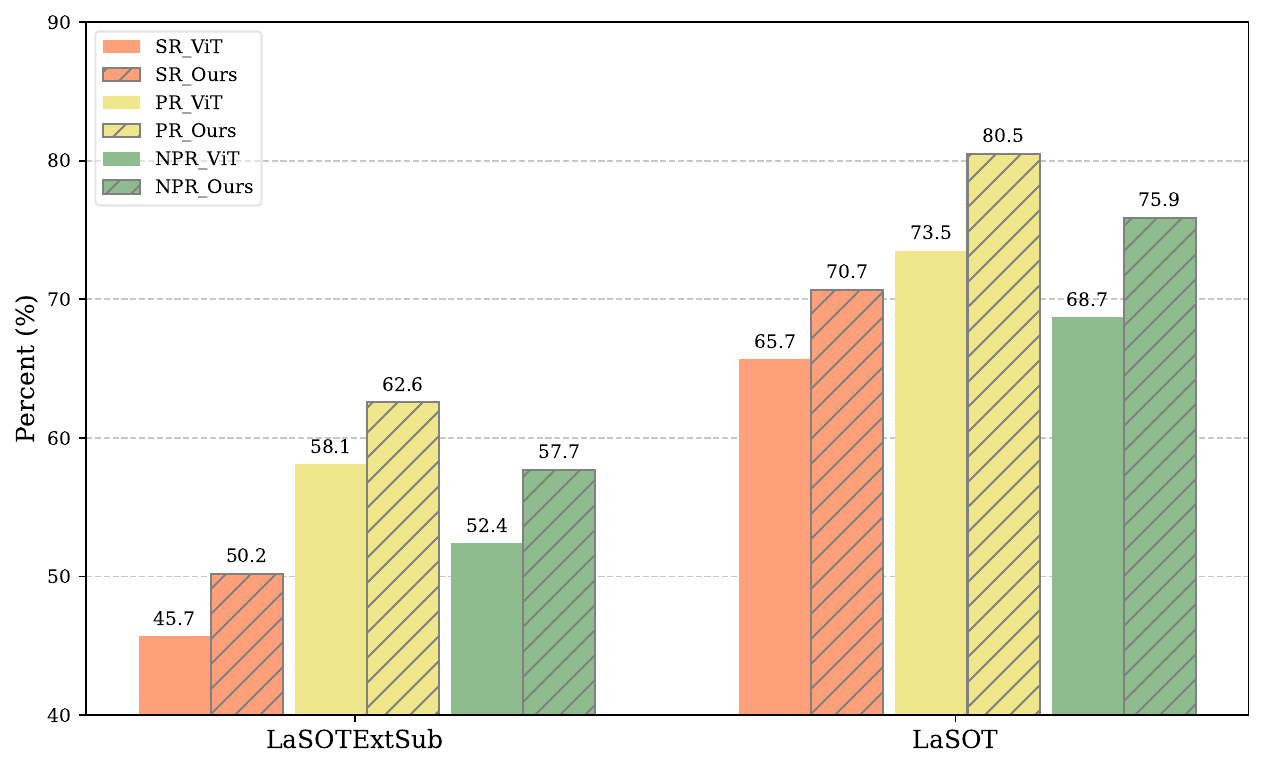} 
\caption{ 
Performance comparison of vanilla ViT \textit{vs.} our tracking-tailored AViT-Enc on the  LaSOTExtSub~\cite{Fan_2020_IJCV_Lasot_ext} and LaSOT~\cite{lasot} datasets. We report success rate (SR), precision rate (PR), and normalized precision rate (NPR). 
Our designed AViT improves performance by around 5\% on each metric on both datasets compared with vanilla ViT.   
 } 
\label{Framework_performance_compare} 
\end{figure}

Recently, it has been pointed out that vanilla ViT fails to exploit image-related prior knowledge, which results in slower convergence and suboptimal performance~\cite{chen2023vision}.
\textcolor{black}{
As demonstrated in prior research~\cite{chen2023vision}, Convolutional Neural Networks (CNNs) have excelled in computer vision tasks by capitalizing on their ability to exploit the spatial structure and local dependencies inherent in images. CNNs exhibit a strong inductive bias for grid-like data and spatial hierarchies. While the Transformer's introduction into the vision field through ViT networks has provided a more universally adaptable capacity for different tasks, it lacks image-specific biases. We argue that the lack of image-specific inductive biases limits the potential of these backbones for visual tracking. 
 }
To address this shortcoming, several hierarchical transformer variants~\cite{liu2021swin,wang2021pyramid} and vision-specific transformers~\cite{xie2021segformer} have been proposed.
\textcolor{black}{Current tracking methods predominantly rely on the general vanilla ViT backbone. In contrast, our primary objective is to devise a tracking-specific ViT backbone to optimize visual tracking performance. To achieve this, we introduce the Adaptor, which incorporates image-related inductive biases tailored for tracking, further improving the full potential of single-branch tracking methods. }
Therefore, we propose an adaptive ViT encoder (AViT-Enc) model prediction network that introduces a specific design in the backbone that leverages spatial priors and vision-specific inductive biases.
To our knowledge, this is the first work to explore the shortcomings of the generic backbone architecture (a vanilla ViT) in visual tracking. 
More specifically, in the encoding phase, we introduce a joint state embedding that encodes the target embedding.
Rather than just providing the target information at the input of the ViT, as done in existing single-branch methods, in our design, we apply an Adaptor that incorporates this information in the ViT branch through layer-wise cross-attention. 
\textcolor{black}{In this way}, our method introduces image-related inductive biases in a single-branch pipeline.
\Cref{Framework_performance_compare} shows that the use of a tracking-based ViT can significantly improve the tracking performance compared to a vanilla ViT design on two datasets.

Another drawback of current trackers is that they fail to consider the temporal consistency of the target position.
In target motion, the position should move approximately consistently, meaning that little position jitter between frames is expected.
This problem is especially urgent when distractor objects are present. 
To address this challenge, we propose an innovative inference approach named CycleTrack, aimed at achieving robust tracking performance in the presence of distractors. 
During the inference phase, CycleTrack incorporates a candidate discrimination module that assesses the reliability of the predicted target candidate. This mechanism examines the temporal cycle consistency between the present frame and the stored previous frame, and then opts for a candidate demonstrating better alignment with this consistency.

To summarize, our contributions are listed as follows:
\begin{itemize}
    \item We propose a tracking-tailored transformer model which exploits image-related inductive biases for single-branch tracking. The method is based on our adaptive ViT encoder which incorporates the target information through layer-wise cross attention. \textcolor{black}{Combined} with the discriminate prediction head, our method unifies the single-branch and discriminate pipelines.
    \item In addition, we propose a novel CycleTrack inference mechanism to enhance the temporal consistency of target location in long-term tracking with little computational overhead and without training costs.
    \item We perform comprehensive experiments to assess the contribution of each element and AViTMP achieves state-of-the-art performance and real-time speed on multiple benchmarks under fair comparison.
\end{itemize}

\section{Related Works}
In this section, we will give a brief review of visual tracking networks and online inference strategies. More related work can be found in the following survey papers~\cite{survey_fahad,survey2}.

\subsection{Visual Object Tracking}
Current popular visual tracking networks can generally be divided into two streams: dual-branch (siamese and discriminative trackers), and single-branch trackers.

In \textbf{dual-branch methods}, 
Siamese tracking paradigm~\cite{siamfc,siamrpn,siamban,li2019siamrpnpp, chen2021transt,stark} conceptualizes tracking as a task involving similarity matching between frames and employs a weight-shared dual-branch backbone. Building upon the contextual interaction and modelling abilities of transformers~\cite{carion2020detr,vaswani2017attention}, Siamese trackers have evolved to incorporate transformers~\cite{chen2021transt,stark,gao2022aiatrack,trdimp} recently.
Discriminative trackers~\cite{kys,atom,prdimp} distinguish the target by minimizing a discriminative objective function. 
KeepTrack~\cite{keeptrack} introduces a learned association network to discern distractors, thereby enhancing tracking robustness. 
Bridging the gap between transformer and discriminative paradigms, ToMP~\cite{tomp} integrates a transformer encoder-decoder module into a concise discriminative target localization model. 
\textcolor{black}{Some works focus on lightweight tracking for efficiency and high-speed running. LightTrack~\cite{yan2021lighttrack} uses neural architecture search (NAS) to build a more lightweight backbone via a one-shot search method. FEAR~\cite{borsuk2022fear} incorporates temporal information with only a single learnable parameter to build a fast Siamese tracker. HiT~\cite{kang2023hit} explores lightweight hierarchical vision transformers which bridge the deep and shallow features for real-time running at edge devices. E.T.Track~\cite{ettrack} introduces a single instance-level attention layer for high-speed running at CPU devices. 
}
Different from the Siamese framework, discriminative approaches have two unbalanced branches, which segregately extract features of training and testing frames.
\textcolor{black}{As mentioned by TransT~\cite{chen2021transt}, the correlation operation (e.g., depthwise correlation) is a simple fusion method to consider the similarity score between the template and search region, which is a local linear matching process. This process will leave similar features but filter different features, especially background information and semantic information, potentially leading to local optimums.
}

\textbf{Single-branch methods}~\cite{chen2022SimTrack,xie2022sbt,gao2023GRM} consolidate feature extraction and interaction ability within a single branch backbone, which naturally avoids the two-branch correlation operation between two branches.
SimTrack~\cite{chen2022SimTrack} simplifies dual-branch feature extraction networks into a unified process and pioneers the introduction of ViT~\cite{visiontransformer2020} into visual tracking.
MixFormer~\cite{mixformer} utilizes a variant of ViT (named CVT~\cite{wu2021cvt}) as its one-branch backbone while splitting the template and search patches for patch embedding at each stage of the backbone.
OSTrack~\cite{ye2022OSTrack} joint feature learning and relation modelling in the one-stream ViT network and integrates a candidate early elimination module after each ViT layer to expedite the inference speed.
SeqTrack~\cite{chen2023seqtrack} employs a vanilla ViT encoder and a causal transformer decoder to locate the target autoregressively.  
GRM~\cite{gao2023GRM} also employs a vanilla ViT but flexibly switches the framework to two-branch or single-branch based on token division, whilst enabling more flexible relation modelling by selecting appropriate search tokens to interact with template tokens.

\textcolor{black}{We} noticed that the current prevailing single-branch trackers commonly borrow the conventional transformer (i.e. VIT~\cite{visiontransformer2020}, Swin~\cite{liu2021swin}) and their variants from the classification task. However, there is no tracker for tailored adaptations of the backbone to suit the unique demands and image-related inductive biases of the visual tracking task.
In this paper, we extend the vanilla ViT encoder specifically tailored for the visual tracking task for the first time.

\begin{figure*}[!t]
\centering
\subfloat[Single-branch tracker]
{\includegraphics[width=0.25\linewidth]{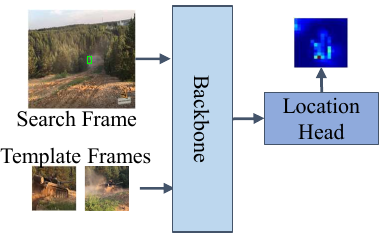}}
\label{overeview_one_stram}
\hfil
\subfloat[Transformer discriminative predictor]
{\includegraphics[width=0.35\linewidth]
{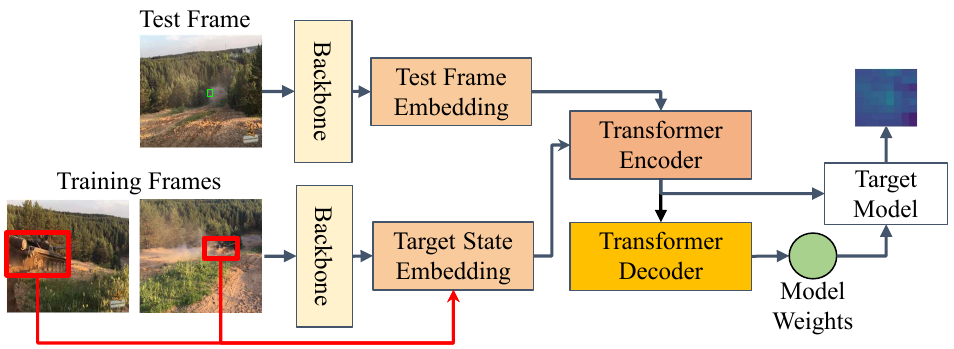} 
}
\hfil
\subfloat[Our AViT discriminative predictor]{\includegraphics[width=0.35\linewidth] {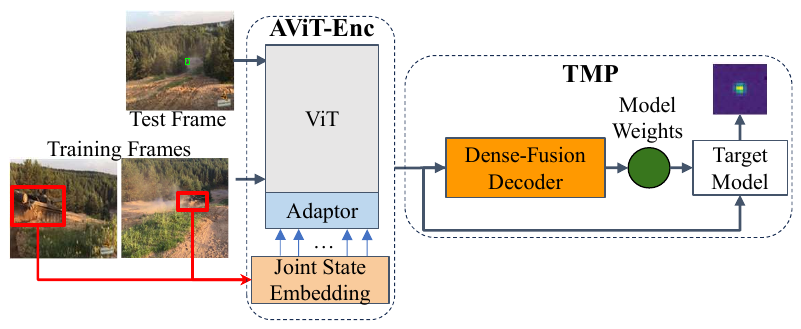}  
} 
\caption{Comparison among trackers that employ (a) single-branch paradigm, (b) discriminative transformer model, and (c) our proposed single-branch discriminative predictor AViTMP. Our model harnesses the strengths of the first two paradigms by incorporating the proposed encoder-decoder design.
In this way, AViTMP integrates the powerful feature extraction capability from single-branch trackers and target prediction ability from discriminate methods into one pipeline. }
\label{fig:overview}
\end{figure*}

\subsection{Online Inference Paradigms}
Discriminative appearance models~\cite{dimp,atom,danelljan2016beyond,henriques2014high} typically incorporate background information during online target classifier learning to boost appearance discriminative capabilities and suppress distractor interference. Nonetheless, appearance models still frequently encounter challenges in effectively distinguishing between distractors and target candidates.
To address this concern, KYS~\cite{kys} extends an RNN on the appearance model to propagate information across frames.
KeepTrack~\cite{keeptrack} proposes an association network with a self-supervised training strategy. However, these solutions introduce extra networks and training costs.
Instead, our AViTMP buffers the previous historic frames and relies on the temporal cycle consistency of the target to discern distractors during inference. This full usage of historical information also contributes to suppressing the tracker's degradation.

On the other hand, template update~\cite{gao2022aiatrack,ltmu,zhang2019learning}, also known as training-frames update, is a widely adopted strategy that fortifies robustness with few computational expenses. It aims to mitigate the limitations of assuming fixed reference templates.
Current existing template updating methods always assume that the initially provided frame serves as the ideal and unalterable template, whereas the second reference (template) requires updating over time.
For example,
STARK~\cite{stark} updates the second template by replacing it when the output fulfils the specified confidence threshold and frame interval criteria.
ToMP~\cite{tomp} updates the second template with a dynamic weight decline and keeps the first template without change.
AiATrack~\cite{gao2022aiatrack} introduces IOUNet~\cite{jiang2018iounet} to get the IOU score and determine whether to update the current frame as a second template and also have a fixed initial template.
Mixformer~\cite{mixformer} appends a trainable network to predict the reliability score as the update condition of the second template.
Consistently, these methods update the second template while maintaining the first template as a fixed reference.
\textcolor{black}{
In contrast, we design a method that reduces dependence on the first reference frame. 
Therefore, we propose a training-free update strategy that pioneers the update of the first reference frame for adapting to the potential violent target change in long-term tracking. }

\section{Method} 
In this section, we introduce an \textbf{A}daptive \textbf{ViT M}odel \textbf{P}rediction method, denoted as AViTMP.
First, we revisit the limitations of the discriminate-based and single-branch trackers in Sec.~\ref{background}.
Subsequently, we provide an overview of our proposed AViTMP in Sec.~\ref{ourmethod}, which proposes a ViT architecture especially adapted for visual tracking.
Further details of the specific design of AViT-Enc and Transformer Model Predictor (TMP) are presented in Sec.~\ref{encoder} and Sec.~\ref{decoder}, respectively. Finally, we detail the online inference pipeline in Sec.~\ref{inference}, which exploits the temporal consistency without training costs.

\subsection{Background} \label{background} 
Two widely recognized paradigms in visual tracking are discriminative model prediction tracking and single-branch transformer tracking.
Single-branch trackers~\cite{chen2022SimTrack,mixformer,xie2022sbt,ye2022OSTrack} commonly integrate template and search frames into one sequence and adopt a robust and efficient backbone for feature extraction.
They then proceed with classification and regression heads to locate the target (~\Cref{fig:overview}a).
However, current single-branch and dual-branch methods all borrow a straightforward backbone (vanilla ViT or Swin) without accounting for tailored design and vision-specific inductive biases for visual tracking.
\textcolor{black}{
On the other hand, dual-branch trackers typically utilize two-branch networks to handle the template and search frames. During the middle stage, the two-branch fusion or discriminative model is essential for predicting future target positions. Among dual-branch methods, discriminative approaches~\cite{dimp,atom,dai2019visual,zheng2020learning} are considered the state-of-the-art pipeline. These approaches involve learning a target model to localize the target object within a test frame, with ToMP (~\Cref{fig:overview}b) being a prominent example.}
Nevertheless, ToMP encodes the training and test frames independently in two branches, which hampers information integration during feature encoding. Furthermore, the divided state encoding procedure between training and test frames inhibits information interaction during the model's initial stages.
In ToMP, another drawback arises from the redundancy of feature encoding components, including a backbone, two state embeddings, a transformer encoder, and a decoder module. While both the backbone and encoder modules are for feature encoding, both embeddings are for feature location.

\begin{figure*}[t] 
\centering 
\includegraphics[width=0.999\textwidth]{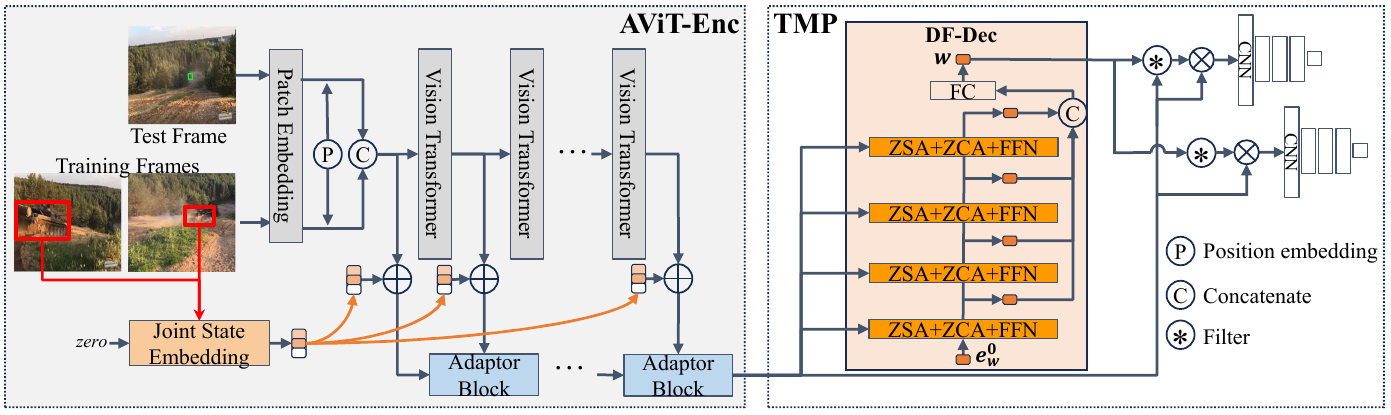}  \vspace{0mm}
\caption{ Overview of our proposed AViTMP architecture, including the AViT-Enc encoder and the Transformer-specific Model Pedictor (TMP) composing of a  DF-Dec decoder and a final target prediction module. 
In AViT-Enc, training and test frames are contacted together and jointly encoded with the target prior information embedding. In DF-Dec, the encoded and adapted features are decoded and then densely integrated layer-wise to generate the model weights. Finally, the target model employs the adapted feature and model weights to predict the target location. 
 } \vspace{0mm}
\label{framework}
\end{figure*}

\subsection{Overview of AViTMP} \label{ourmethod}
To solve the above limitations in both single-branch and discriminate prediction trackers, we present an innovative tracking-tailored target model to optimize these two methods (as depicted in ~\Cref{fig:overview}c).
\textcolor{black}{In the current discriminate tracking method, the decoding process is generally divided into two branches, and each branch models its own training or test frames independently. This method ignores the connection between different frames, while our method concatenates the two branches together and uses a joint embedding module to build the connection between training and test frames. Compared with individual decoding, our joint decoding can simplify the two decoding operations into one decoding process. This single-branch decoding pipeline can naturally contribute to the target model to find more distinctive features with the joint decoding method. }
By doing so, our encoding model becomes adept at incorporating the target state and specific priors right from the outset. This enables subsequent components like the decoder and target model to concentrate more on the target's distinctive features. Additionally, in contrast to extracting a fixed feature space for test frames, this collaborative encoding process can dynamically construct an adaptive feature space connected to the training frames for every test frame.

The pipeline of our tracker is depicted in ~\Cref{fig:overview}c. Similar to discriminative trackers, the input consists of both test and training frames. 
\textcolor{black}{Different from two-branch methods, in our encoding phase, we propose the specifically designed adaptive module and the joint state embedding. Further, instead of separating the extraction and encoding of features into distinct stages and branches as is done in two-branch methods, AViTMP facilitates the joint decoding of discriminative features from both test and training frames through a single-branch network. With this pipeline, our encoding model becomes adept at incorporating the target state and specific priors right from the outset, while two-branch methods correlate target features in the middle stage.}
Initially, we perform a joint encoding of these frames using the proposed adaptive VIT encoder.
The joint state embedding integrates target position priors into the extracted features to augment the target region. Subsequently, these encoded features are directed to the dense-fusion decoder for predicting both the model weights and the target model. Ultimately, the target model discriminates against the target by considering the model weights and the encoded features.

\subsection{AViT-Enc: Adaptive ViT Encoder}  \label{encoder}
Other than standard single-branch tracking methods~\cite{chen2022SimTrack,chen2023seqtrack,ye2022OSTrack}, our Adaptive ViT Encoder (AViT-Enc) aims to explicitly incorporate image-related inductive biases through the tailored proposed Adaptor and state embedding module. An overview of AViT-Enc design is shown in ~\Cref{framework}.

With the input test frame $x_{test} \in \mathbb{R}^{H\times W\times 3} $ and training frames $x_{i} \in \mathbb{R}^{H\times W\times 3} $, the joint encoding function is:
\begin{equation}
    \textcolor{black}{\mathcal{F}^{L}_{avit}} = \mathbf{AViT}\textit{-}\mathbf{Enc}([x_{1},...,x_{m},x_{test}]) \quad  
\end{equation}
AViT-Enc consists of four modules: Patch Embedding ($\mathbf{PE}$), Vision Transformer layers ($\mathbf{ViT}$), Joint State Embedding ($\mathbf{JSE}$) and Adaptor blocks ($\mathbf{Adaptor}$). 
The input frames are firstly flattened and projected to C-dimensional tokens by a patch embedding block. Then, they are concatenated together into a patch sequence and then add position embedding to get $\mathcal{F}^0_{vit} \in \mathbb{R}^{h\times w \times C}$.
Here, $h=\lceil H/p \rceil$ and $w=\lceil W/p \rceil$ are \textcolor{black}{hight and weight} of each patch.
Therefore, each frame is divided into $p\times p$ non-overlapping patches. For position embedding, the same learnable position embeddings~\cite{vaswani2017attention} $\mathbf{Pos}\in \mathcal{R}^{p*p}$ are \textcolor{black}{added to} each test and training frame, formulated as:
\begin{equation}
\begin{split}
 \mathcal{F}^{0}_{vit} = [& \mathbf{PE}(x_{test})+\mathbf{Pos}, \mathbf{PE}(x_{1})+\mathbf{Pos},\\
 &\quad \quad \quad ...,\quad \quad \quad \,\,\, \mathbf{PE}(x_{m})+\mathbf{Pos}]
\end{split}
\end{equation}
Subsequent AViT blocks maintain a consistent spatial scale between the input and output. There are $L$ encoder layers in total.
We embed an Adaptor module inside to adapt the vanilla ViT for tracking tasks, as shown in ~\Cref{framework}.

Concretely, our approach involves a step-wise process.
Initially, the first layer of AViT is directly embedded from $\mathcal{F}^{0}_{vit}$ while the following $j$-th AViT feature $\mathcal{F}^{j}_{avit}$ is obtained from the last Adaptor layer.
In the context of the $j$-th layer feature of ViT denoted as $\mathcal{F}^{j}_{vit}$, the next layer feature $\mathcal{F}^{j+1}_{vit}$ is extracted by $\mathbf{ViT}_{j+1}$.
Subsequently, the next AViT layer features $\mathcal{F}^{j+1}_{avit}$ is built by Adaptor.
This process can be formulated as follows:
\begin{equation}
\begin{split}
      & \mathcal{F}^{0}_{avit}=\mathbf{JSE}(\mathcal{F}^{0}_{vit}), \\ 
      & \mathcal{F}^{j+1}_{vit} = \mathrm{\mathbf{ViT}_{j+1}}(\mathcal{F}^{j}_{vit}), \quad  \mathcal{\hat{F}}^{j+1}_{vit} = \mathbf{JSE}(\mathcal{F}^{j+1}_{vit}), \\
      & \mathcal{F}^{j+1}_{avit}= \mathbf{Adaptor}_j(\mathcal{F}^{j}_{avit},  \mathcal{\hat{F}}^{j+1}_{vit})
\end{split}
\end{equation}
where $j\in [0,...,L-1]$. With this layer-wise $L$ feature interaction, we obtain the final adaptive feature $\mathcal{F}^{L}_{avit}$. 
\textcolor{black}{Our \textbf{JSE}} integrates target location information from both the test and training frames into the encoder feature.

\minisection{Joint State Embedding.}
To leverage the target prior and spatial biases, we introduce a joint state embedding module (\textbf{JSE}) designed to incorporate foreground center knowledge and bounding box edge position information into the extracted features.
Particularly, for training frames, we employ the learnable embedding $e_\mathrm{fg} \in\mathbb{R}^{1\times C}$ to represent their foreground, \textcolor{black}{and} the Gaussian center label $y_i\in\mathbb{R}^{h\times w \times 1}$ to introduce the target center inductive bias. Inspired by ToMP~\cite{tomp}, we also introduce the target bounding box prior $d_i$ with a multi-layer perceptron $\phi$ to highlight the bounding box edge and to strengthen the location ability, formulated as:
\begin{equation}
\begin{split}
        & \psi_i = y_i \cdot e_\mathrm{fg} ,  \quad \quad \quad \phi_i = \textcolor{black}{\textbf{FC}}(d_i), \\
        & \psi_{test} = \textit{zero}\cdot e_\mathrm{fg}, \quad  \phi_{test} = \textcolor{black}{\textbf{FC}}(\textit{zero}), \\
      &\mathcal{\hat{F}}^{j+1}_{vit} = \mathcal{F}^{j+1}_{vit}+[\psi_0,...,\psi_m,\psi_{test}] + [\phi_0,...,\phi_m,\phi_{test}]
\end{split}
\end{equation}
\textcolor{black}{\textbf{FC} is a Fully-Connected layer.} 
For each ViT block output, our joint location embedding is weight-shared to get the same target state embedding for different feature layers.
\textcolor{black}{Due} to the test frame prior information is not available during inference, in the joint location embedding process, we set the test frame location embedding as \textit{zero} to maintain consistency in the training and inference phases.
Besides, existing single-branch methods ignore relevant background information by only considering the target cropped region as a template. Instead, we consider the whole source square frame as the training frame. \textcolor{black}{In detail}, we extract the whole foreground and background features and highlight the foreground via the feature embedding of the target state spatial prior.

\minisection{Adaptor.}
In our Adaptor module, each block is built by zero-center cross-attention (ZCA) and a feed-forward network (FFN) with a residual connection. 
This process calculates the cross-attention of different feature spaces between test and training frames, further integrating the target state prior and inductive biases for tracking.
Different from existing trackers which need \textcolor{black}{sinusoidal} positional encoding added before each attention block, in our method, we remove the position encoding of each block in all attention calculation processes. That means we only append \textcolor{black}{learnable} position embeddings to each patch at the very beginning of our tracker (in ViT layers) to avoid redundant position embedding operations for different modules.
$\mathbf{Adaptor}_j$ layer is formulated as:
\begin{equation} 
\begin{aligned}
&\mathcal{\hat{F}}^{j+1}_{avit}= \mathcal{F}^{j}_{avit}+\mathbf{ZCA}_j \left(\mathcal{F}^{j}_{avit}, \mathcal{\hat{F}}^{j+1}_{vit},  \mathcal{\hat{F}}^{j+1}_{vit}\right), \\
&\mathcal{F}^{j+1}_{avit} = \mathcal{\hat{F}}^{j+1}_{avit}+ \mathbf{FFN}_j(\mathcal{\hat{F}}^{j+1}_{avit})
\end{aligned}
\end{equation}
In each layer, zero-center attention block is formulated as:
\begin{equation}
\mathbf{ZCA}(\mathbf{Q},\mathbf{K},\mathbf{V}) 
 =softmax(\frac{\mathbf{(Q-\mu_q)(K-\mu_k)}^{T}}{\sqrt{d_{k}}}) \mathbf{V}
\label{eq.zca}
\end{equation}
where $\mu$ means the average of each vector. $\mu_q=\frac{1}{N}\sum_{j=1}^N{q_j}$, $\mu_k=\frac{1}{N}\sum_{j=1}^N{k_j}$, $N$ indicates the number of attention block heads\footnote{\textcolor{black}{Note that the number of heads is set to eight for all attention modules in our network.}}.

\textcolor{black}{
In contrast to the conventional attention mechanism, in ZCA the average of Q and K is subtracted during attention calculation. 
This keeps the query and key vector values both within the range of (0, 1) to avoid over-smoothing in deep layers. 
}
It is inspired by the claim: that the attention mechanism inherently amounts to a low-pass filter, and the stack of multi-head attention layers in the transformer may suppress the Alternating Current (AC) component of features severely and only leave the over-smoothing Direct Current (DC) component~\cite{nt2019revisiting, cai2020note,tang2023learning}.
Therefore, we remove the DC component by subtracting the mean value of $\textbf{Q}$ and $\textbf{K}$ to leave the AC component for optimization.
This can be conceptualized as a process of eliminating the DC information while retaining the AC components of the input features.
Naturally, in Zero-centered Self-Attention (ZSA), we set \textcolor{black}{$\textbf{Q}=\textbf{K}$} in Eq.~\ref{eq.zca}.

Finally, we obtained the AViT-Enc feature $ \mathcal{F}^{L}_{avit} $ from the last Adaptor block. 
Our Adaptor has been partially motivated by Chen et al.~\cite{chen2023vision}  which incorporate image-related inductive biases into the transformer design. \textcolor{black}{Chen’s ViT adapter introduces three heavy components, consisting of a spatial prior module (consisting of a ResNet-block, 3×3 Conv layers, and 1×1 Conv layers), a spatial feature injector (consisting of a cross-attention), and a multi-scale feature extractor (consisting of a cross-attention and an FFN). Compared with Chen’s ViT adapter, ours, however, has significantly fewer network blocks and parameters. Our Adaptor contains only a cross-attention and FFN layer, which is a lightweight component to guarantee real-time tracking with only a few additional parameters.  
}

\subsection{Transformer Model Predictor} \label{decoder}

\minisection{Dense-Fusion Decoder.}
In our dense-fusion decoder (DF-Dec), we feed $ \mathcal{F}^{L}_{avit} $ as the input.
We initialize a learnable embedding $e^0_\mathrm{w} \in\mathbb{R}^{1\times C}$ as the query of transformer blocks to predict the target model weight. As shown in ~\Cref{framework}, \textcolor{black}{DF-Dec consists of six transformer decoder layers \textcolor{black}{with a FC} layer.} 
\textcolor{black}{Each transformer layer is built by a zero-center self-attention (ZSA), a zero-center cross-attention (ZCA), and an FFN block.} 
Position encoding of each attention block is also removed.
Each layer can be written as:
\begin{equation}
\begin{aligned}
&\mathcal{\hat{F}}^{L}_{avit}=\mathcal{F}^{L}_{avit}+\mathbf{ZSA}_i(\mathcal{F}^{L}_{avit}, \mathcal{F}^{L}_{avit}, \mathcal{F}^{L}_{avit}), \\
&\hat{e}^{i-1}_\mathrm{w}=e^{i-1}_\mathrm{w}+\mathbf{ZCA}_i(e^{i-1}_\mathrm{w}, \mathcal{\hat{F}}^{L}_{avit}, \mathcal{\hat{F}}^{L}_{avit} ), \\
&e^i_\mathrm{w}=\hat{e}^{i-1}_\mathrm{w}+\mathbf{FFN}_i(\hat{e}^{i-1}_\mathrm{w}) \\
\end{aligned}
\end{equation}
where $i \in [1,...,6]$. Then, we concatenate each layer feature and project it 
as the target model weights:
\begin{equation}
\begin{aligned}
&w= LN(\mathbf{FC}([e^1_\mathrm{w},...,e^6_\mathrm{w}])) \\
\end{aligned}
\end{equation}
where $LN$ denotes layer normalization.

\minisection{Target Model.}
In the target model, we set a residual connection way of the input feature $ \mathcal{F}^{L}_{avit} $. 
Following discriminative target model methods~\cite{dimp,tomp}, we generate the target scores by the model weights and encoded features, formulated as:
\begin{equation}
\begin{aligned}
h(w, \mathcal{F}^{L}_{avit}) = w \ast \mathcal{F}^{L}_{avit}
\end{aligned}
\end{equation}
where $w \in\mathbb{R}^{1\times C}$ are the weights of the convolution filter.

\minisection{Target Location.} 
\textcolor{black}{In the prediction heads, we employ two same parallel networks with non-sharing weights.}  
\textcolor{black}{In the regression head, we firstly} adopt an FC layer to obtain the weights of regression $w_{reg}$ based on the target model weights $w$. Then, we use the filter to compute the attention weights and multiply the attention weights point-wise with features. Finally, the weighted features are fed into a CNN network to regress the bounding box edge $\hat{d}$, format as 
\begin{equation}
\begin{aligned}
    & w_{reg} = \mathbf{FC}_1(w), \\
    & w^{reg}_{attn} = h(w_{reg}, \mathcal{F}^{L}_{avit}), \\
    & \hat{d} = \mathbf{CNN} (w^{reg}_{attn} *  \mathcal{F}^{L}_{avit})
\end{aligned}
\end{equation} 
where $\mathbf{CNN}$ consists of five convolutional layers. 
For classifier, instead of previously discriminative classifiers~\cite{dimp,atom,tomp} which directly use the target model $ h(w, \mathcal{F}^{L}_{avit})$ as the output, 
we keep the same procedure with the regression process except the output CNN dimension is set to 1.
This design tailors the model to focus on foreground discrimination, which balances the predictions from the two heads.

\begin{figure}[t]
\centering 
\includegraphics[width=0.99\columnwidth]{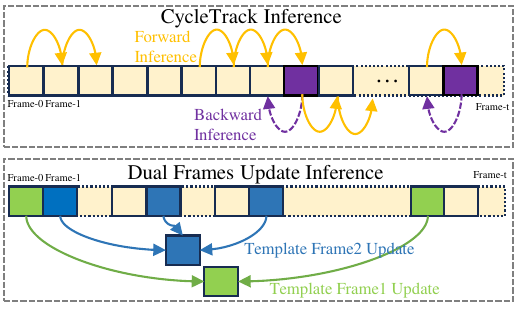} 
\vspace{0mm}
\caption{
\textcolor{black}{Overview of proposed online inference strategies. (1) CycleTrack (top) consists of two different tracking processes, named forward and backward inference; (2) Dual-Frames Update (bottom) update two templates over time and conditions.
 } }
\label{inference_strategies} 
\end{figure}

\subsection{Online Inference Strategies}  \label{inference}
During inference, we implement two strategies to bolster tracking robustness without incurring any additional training expenses and with minimal extra inference overhead. 
\textcolor{black}{
As shown in \Cref{inference_strategies}, our CycleTrack consists of forward inference and backward inference,  two different processes. Forward inference is the same with the current tracking pipeline while backward inference will be active (purple box) when the prediction is not reliable enough. The backward inference uses the current frame as prior and predicts the bbox in the previous frame to check the prediction's credibility. 
Dual-Frames Update briefly shows how to update two templates over time. During inference, the first frame and second frame are separate as template1 and template2, while our method will update these two templates independently. This dual-frame update method can make the tracking more robust in the long-term tracking process. 
}

\begin{figure}[t]
\centering 
\includegraphics[width=0.99\columnwidth]{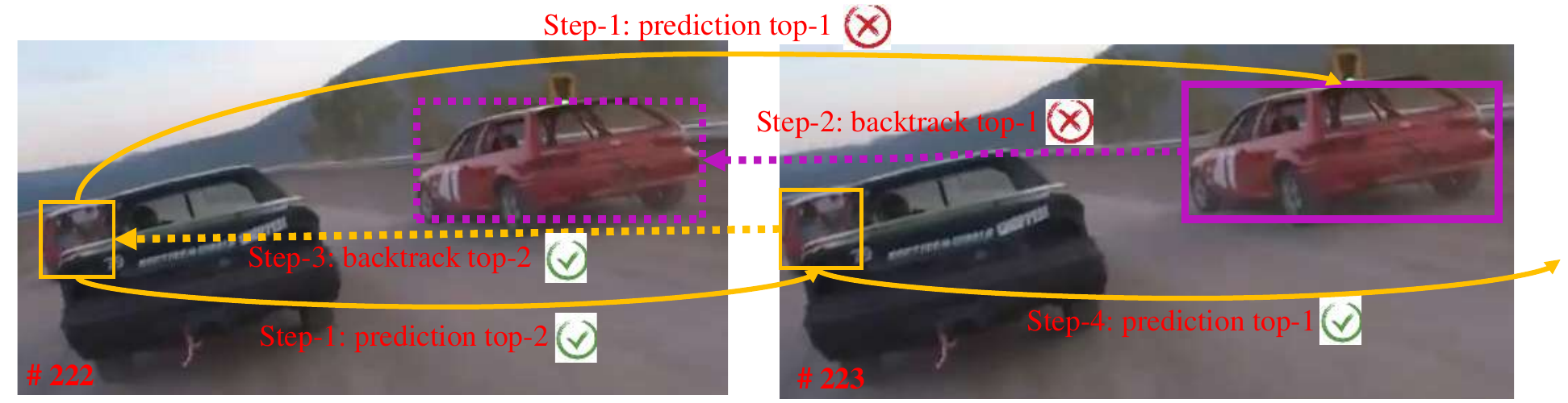} 
\vspace{0mm}
\caption{\textcolor{black}{Spatial consistent and inconsistent prediction along the temporal adjacent frames. The solid line represents the forward track process, while the dotted line indicates the backward track process to verify the prediction of top-2 results.
 } }
\label{cycletrack_rule} 
\end{figure}

\minisection{CycleTrack Pipeline.}
During the inference stage, \textcolor{black}{all previous} pipeline progresses frame by frame in chronological order. 
While this adheres to the temporal sequence, it encounters occasional failures or mis-tracks, particularly in the presence of distractors. 
Differently, we propose a CycleTrack inference pipeline that rectifies the predicted candidate in a backward direction tracking when temporal cycle consistency is not satisfied, as shown in ~\Cref{inference_strategies}.
\textcolor{black}{CycleTrack hinges on the principle based on “temporal cycle consistency” which means the spatial position keeps consistency along the temporal frames. It means the bounding boxes in two nearby frames should be relatively close to each other rather than having drastic position jumps. However, current visual tracking only considers the top-one prediction bounding box as the final result, and do not consider the prediction consistency. As shown in ~\Cref{cycletrack_rule}, the current inference method only chooses the top-1 box as the final result. However, the purple prediction bbox has a dramatic spatial jump compared to the orange bbox in the previous frame, which does not obey the spatial consistency rule. Based on this, we activate CycleTrack to discriminate which box is the more reliable one in the top two bounding boxes. }
This signifies that a precise target candidate, when retrogressively tracked through time, should ultimately return to the localization of the current bounding box. The underlying hypothesis is that candidates demonstrating spatial position consistency in nearby frames are more likely to be the accurate target. In instances where there are multiple potential target candidates, CycleTrack acts as a criterion for target selection.

Our target and distractor discrimination process is formulated in Algorithm~\ref{algorithm1}.
Concretely, given an initial frame $\mathbf{I}^{t-1}$ and the target prediction model $\mathbf{\Psi}$, \textcolor{black}{ we estimate the target candidate bounding box $\textbf{B}^{t}_1$ in frame $\mathbf{I}^{t}$ based on the previous frame $\mathbf{I}^{t}$. Then, we mask the prediction box region $\textbf{B}^{t}_1$ in $\mathbf{I}^{t}$ to extract the distractor candidate $\textbf{B}^{t}_2$ with the sub-top response. } 
Then, we use the threshold $\tau_c$ as a quality measurement for the prediction score of $\textbf{B}^{t}_1$. When the classification score $\mathbf{S}^{t}_1$ is lower than $\tau_c$, \textcolor{black}{backward track} will be activated to choose a more reliable box.
Next, we employ $\mathbf{I}^{t}$ as the previous test frame and consider $\mathbf{I}^{t-1}$ as the future test frame to achieve the backward track.
Based on the center and scale of $\textbf{B}^{t}_1$ in $\mathbf{I}^{t}$, we resize the frame $\mathbf{I}^{t-1}$ accordingly and get the model-predicted candidate output $\mathbf{\hat{B}}^{t-1}_1$. \textcolor{black}{ 
 Similarily, based on $\textbf{B}^{t}_2$, we predict the box $\mathbf{\hat{B}}^{t-1}_2$.
By comparing the backward-track results $\mathbf{\hat{B}}^{t-1}_1$ and $\mathbf{\hat{B}}^{t-1}_2$ with the ground-truth $\mathbf{B}^{t-1}_2$, we choose the more successful one to correct the prediction result $\textbf{B}^{t}_1$. }

\begin{algorithm}[t]
    \caption{ \textcolor{black}{ CycleTrack Inference Pipeline.} }
    \label{algorithm1} 
    \textbf{Input:} Sequence $\mathcal{I}=[\mathbf{I}^1,...,\mathbf{I}^n]$,  target predictor $\mathbf{\Psi}$, threshold $\tau_c=0.5$ \\
    \textbf{For} $t=1,\dots,n$ \textbf{do}
    \begin{algorithmic}
        \STATE  \textcolor{gray}{\# get the top and sub-top predicted bbox $\mathbf{B}$ and prediction score $\mathbf{S}$. }
        \STATE  $ (\mathbf{B}^t_1,\mathbf{S}^{t}_1), (\mathbf{B}^t_2,\mathbf{S}^{t}_2) \gets  \mathbf{\Psi}(\mathbf{I}^t, \mathbf{B}^{t-1}_1)$  \hfill \textcolor{gray}{$\triangleright$  Forward track}
        \STATE \textbf{IF} $\mathbf{S}^{t}_1 < \tau_c $  \textbf{then}  \hfill  \textcolor{gray}{ $\triangleright$ Activate BackTrack at $\mathbf{I}^t$ }
        \STATE ~~~~$(\mathbf{\hat{B}}^{t-1}_1,\mathbf{\hat{S}}^{t-1}_1) \gets \mathbf{\Psi}(\mathbf{I}^{t-1}, \mathbf{B}^{t}_1)$   \hfill  \textcolor{gray}{ $\triangleright$Back track on $\mathbf{B}^{t}_1$}
        \STATE ~~~~$(\mathbf{\hat{B}}^{t-1}_2,\mathbf{\hat{S}}^{t-1}_2) \gets \mathbf{\Psi}(\mathbf{I}^{t-1}, \mathbf{B}^{t}_2)$   \hfill  \textcolor{gray}{ $\triangleright$ Back track on $\mathbf{B}^{t}_2$}
        \STATE ~~~~\textbf{IF} $ \text{IOU}(\mathbf{\hat{B}}^{t-1}_2,\mathbf{B}^{t-1}_1)>\text{IOU}(\mathbf{\hat{B}}^{t-1}_1,\mathbf{B}^{t-1}_1) $    
        \STATE ~~~~\& $\mathbf{\hat{S}}^{t-1}_2 > \tau_c$  \textbf{then} \hfill   \textcolor{gray}{ $\triangleright$ Correct the error prediction }
        \STATE ~~~~~~~~~$ (\mathbf{B}^t_1,\mathbf{S}^{t}_1) = (\mathbf{B}^t_2, \mathbf{S}^{t}_2)$   \hfill   \textcolor{gray}{ $\triangleright$ Correct the prediction }
    \end{algorithmic} 
    \textbf{Output:} $\mathcal{B}=[\mathbf{B}_1^1, \dots, \mathbf{B}_1^n]$
\end{algorithm}

\minisection{Dual-Frames Update.}
In AViTMP training, we employ two training frames and one test frame for joint encoding. \textcolor{black}{The training frames are also called template frames.} Therefore, we will search for two training frames that contain the bounding box as a reference to find the target location in the test frame. 
\textcolor{black}{ In the beginning, our approach involves using the initial frame with an annotated bounding box as the first training frame. The second training frame is initialized as a zero vector and then updated following the strategy introduced in ToMP~\cite{tomp}. } 
Specifically, the second training frame is updated with the most recent frame when its classifier score surpasses a predefined threshold $\tau_2$.
However, unlike existing methods that overlook the need for updating the first training frame in response to significant changes in the target's scale, our approach addresses this issue. We incorporate updates to the initial training frame when detecting substantial scale changes using a high confidence threshold denoted as $\tau_1$.

\input{tables/lasot}

\section{Experiments} 
\subsection{Implementation Details} \label{training}
Our method AViTMP is implemented using PyTracking framework~\cite{Danelljan_2019_github_pytracking}. The training dataset incorporates COCO~\cite{coco}, LaSOT~\cite{lasot}, GOT10k~\cite{got10k}, and TrackingNet~\cite{trackingnet}.
The training regimen comprises 300 epochs, encompassing the sampling of 40,000 sub-sequences.
In the mini-batch training procedure, we select $m=2$ training frames and a single test frame from a video sequence, arranging them chronologically ($x_1<x_2<x_{test}$). Consistent with recent discriminate models, uniform resolution is maintained for both test and training frames. Specifically, all three frames are resized to dimensions of $288\times288\times3$. 
The patch embedding layer within ViT comprises a convolutional block, projecting frames into patch sequences of size $18\times18$.
Subsequent encoding operations ensure a consistent dimensionality between input and output features. Consequently, both the decoder and the head sections maintain a default value of $C$=768.
Regarding the vanilla ViT network, we initialize the pre-trained model using the unsupervised model MAE~\cite{he2022mae}, while the remaining modules are trained from the ground up. The learning rate undergoes decay by a factor of 0.2 after 150 and 250 epochs.
The optimization process is carried out using the AdamW optimizer~\cite{adamw}, facilitated by 4 NVIDIA A40 GPUs.
To ensure the consistency of training state-prior information and the uniformity of prior distribution throughout training and testing, center and scale jitters are deliberately excluded during the training phase.  Finally, the inference speed of AViTMP is 40 FPS (frame per second) on an A40 GPU. 
\textcolor{black}{All our results are averaged by 3 times running. }

\subsection{Network Setting}
\minisection{Training.}
For the classification head, we adopt the LBHinge~\cite{dimp} loss following DiMP~\cite{dimp}. In the regression head, we only adopt GIOU~\cite{Rezatofighi_2019_CVPR_GIOU} loss to converge the four edges of the predicted box. Note that we do not use the popular $L1$ loss in single-branch trackers since the full usage of the bounding box prior information \textcolor{black}{contributes enough to the network convergence} in the joint encoding procedure.
The final loss is:
\begin{equation}
\begin{aligned}
L_{final} = \lambda L_{cls}(\hat{y}, y) + L_{giou}(\hat{d}, d)
\end{aligned}
\end{equation}
Here, $\lambda=200$ is the weight to increase the classification loss magnitude for training. 
$y, d$ denote the Gaussian center label and ground truth prior, respectively.

\minisection{Inference.}
\textcolor{black}{
$\tau_c$ is a hyperparameter that is set to 0.5 in our paper. It is the threshold to determine whether to activate CycleTrack. This means that when the prediction accuracy is lower than 0.5, the prediction is more likely wrong, we activate backward track to check the top 2 boxes for robust tracking. Besides, $\tau_c$ also serves as the threshold for rectifying erroneous output results stemming from the network predictions.  We use the same hyperparameter on all datasets and experiments.
}
Regarding the two training-frames update, the threshold for updating the second template is set as $\tau_2$=0.85.
Concerning the initial training frames, updates are triggered under two conditions: if the predicted classifier's confidence surpasses $\tau_1=1.00$, and if the target size experiences a dramatic change (exceeding 16 times) in comparison to the initial training frame (in terms of zooming in or out).
\textcolor{black}{As outlined in Sec.~\ref{decoder}, discriminate trackers are different from other classification-regression prediction trackers which use the prediction confidence score directly as the classification score. In our discriminate tracking head, we utilize a convolution kernel to multiply the prediction score (which$<=$1.00) with the model weight (possibly$>$1.00). This model weight enables the classification output to represent the weighted confidence score, which may be larger than 1.0. Consequently, setting the threshold $ \tau_1$  set as 1.00 is justified. It will activate when the weighted confidence score is larger than 1.00 (generally when the model weight $>$1.00  while the prediction score is between 0.9 and 1.00).  This mechanism allows for updating the first frame when the network assigns a high weight and a very confident prediction score with the threshold $\tau_1$  as 1.00. We set it as 1.00 to avoid harmful updates involving distractors during the training frames update process to improve the robustness performance. }

\subsection{Comparison to the State of the Art}
In this section, we evaluate our proposed tracker on eight long-time, large-scale benchmarks, including LaSOT, LaSOTExtSub, AVisT, VOT2020\_Bbox, UAV123, TNL2k, TrackingNet, and VOT2020\_Mask. We report our tracking performance with current state-of-the-art trackers with a fair comparison condition.
Currently, most trackers use the base version of ViT (ViT-Base) for a fair comparison. In contrast, some trackers present performance-oriented variants by using large version backbones (i.e., ViT-Large) and large resolution (i.e., 384). In this section, we compared the performance with the same base version of the backbone (ViT-Base) for a fair comparison of the convincing module design.

\input{tables/lasotext}

\input{tables/avist}

\input{tables/vot2020}

\minisection{Results on LaSOT~\cite{lasot}:} LaSOT is a large-scale long-term dataset composed of 280 test videos with 2500 frames on average. ~\Cref{tab:lasot} presents the evaluations of trackers in terms of area-under-the-curve (AUC), precision, and normalized precision. Our method AViTMP showcases superior performance over recent discriminative trackers such as ToMP~\cite{tomp} and KeepTrack~\cite{keeptrack}, boasting a substantial performance gap. It is worth highlighting that, in comparison to contemporary vanilla ViT single-branch approaches SeqTrack-B256~\cite{chen2023seqtrack} and OSTrack256~\cite{ye2022OSTrack}, AViTMP outperforms them by achieving a new state-of-the-art performance (AUC of 70.7\%).

\minisection{Results on LaSOTExtSub~\cite{Fan_2020_IJCV_Lasot_ext}:} LaSOTExtSub is an extension dataset of LaSOT. It contains 15 new classes with 150 test sequences in total. LaSOTExtSub also encompasses lots of long-term sequences with distractor scenarios.
As shown in ~\Cref{tab:lasotext}, under the conditions of aligned settings, AViTMP attains a new state-of-the-art performance at 50.2\% AUC, surpassing the discriminative model ToMP101 by a significant margin of 4.3\%. When considering an equivalent vanilla ViT-B backbone and similar resolution, AViTMP also outperforms state-of-the-art single-branch tracker SeqTrack-B256 by 0.7\% in terms of AUC.

\minisection{Results on AVisT~\cite{noman2022avist}:} AVisT is a recently released benchmark that comprises 120 sequences in a variety of adverse scenarios highly relevant to real-world applications, such as bad weather conditions and camouflage.
As shown in ~\Cref{tab:avist}, compared with the trackers of similar complexity, our tracker outperforms the other trackers, including GRM and MixFormer-22k, setting a new state-of-the-art AUC score under challenging scenarios.

\minisection{Results on VOT2020\_Bbox~\cite{Kristan_2020_ECCVW_VOT2020}:} VOT2020 contains 60 videos, and we compare the top methods in VOT challenge~\cite{Kristan_2020_ECCVW_VOT2020}. Instead of the one-pass evaluation, the trackers are evaluated following the multi-start protocol which is specifically suited for the VOT challenge.
Since AViTMP is end-to-end supervised by bounding boxes solely, we compare the bounding box trackers in ~\Cref{tab:vot2020}. 
Our approach performs better in terms of overall performance (EAO) compared with the Siamese method CSWinTT, discriminate method ToMP101 and single-branch method SeqTrack-B256. Especially, AViTMP achieves 0.840 in the robustness aspect, outperforming SeqTrack-B256 and ToMP101 with 3.4\% and 2.6\% respectively.
In the multi-start protocol, the quality of the initial frame is hard to predict, making the evaluation results much closer to a real application. The robustness metric (represents tracking failure times) comparison shows the powerful effectiveness of our inference strategies in contributing to robustness tracking.

\input{tables/uav_tnl2k}

\input{tables/lasot_attributes}

\minisection{Results on UAV123~\cite{Mueller_2016_ECCV_UAV123}:} UAV123 is a dataset with 123 test videos captured from UAVs, mainly containing fast motion, small targets, and distractors. It is over 1200 frames on average in a video. ~\Cref{tab:uav_tnl} shows that AViTMP achieves a 70.1\% AUC score, outperforming the previous distractor inhibition two-branch discriminate method KeepTrack and one-branch method SeqTrack.

\minisection{Results on TNL2k~\cite{tnl2k}:} TNL2k is a large-scale and newly created dataset with 700 sequences. ~\Cref{tab:uav_tnl} reports the results following the bounding-box guided tracking rule and AViTMP achieves 54.5\% AUC, competitive with SeqTrack-B256 and OSTrack-256.

\minisection{Results on TrackingNet.}
TrackingNet~\cite{trackingnet} encompasses a collection of 511 test videos. In contrast to the above-mentioned long-term datasets, TrackingNet is positioned as a short-term dataset (around 500 frames for each video) and is relatively less challenging in terms of long-term tracking attributes such as distractors and out-of-view scenarios. The outcomes presented in ~\Cref{trackingnettable} indicate that our tracker achieves a suboptimal AUC of 82.8\%.

\minisection{Results on VOT2020\_Mask.}
In contrast to previous VOT challenges~\cite{Matej_2018_ECCVW_VOT2018, Kristan_2019_ICCVW_VOT2019}, where sequences in the VOT challenge were annotated with bounding boxes, the VOT2020 challenge also incorporates evaluation method based on segmentation masks in each frame. We additionally evaluated our method in VOT2020 by incorporating the segmentation model HQ-SAM~\cite{ke2023samhq}.
As highlighted in ~\Cref{tab:vot2020mask}, our tracker attains an EAO of 0.504, coupled with a robustness performance of 0.821.
With the tracking combined with the segmentation pipeline, segmentation mask accuracy heavily depends on the segmentation quality, and tracking robustness mainly depends on our tracking method. The comparison outcomes further underscore the robust tracking ability of AViTMP.

\input{tables/trackingnet} 

\input{tables/vot2020_mask}

\subsection{Visualization and Analysis}
\begin{figure} 
\centering   \vspace{0mm}
\includegraphics[width=1\columnwidth]{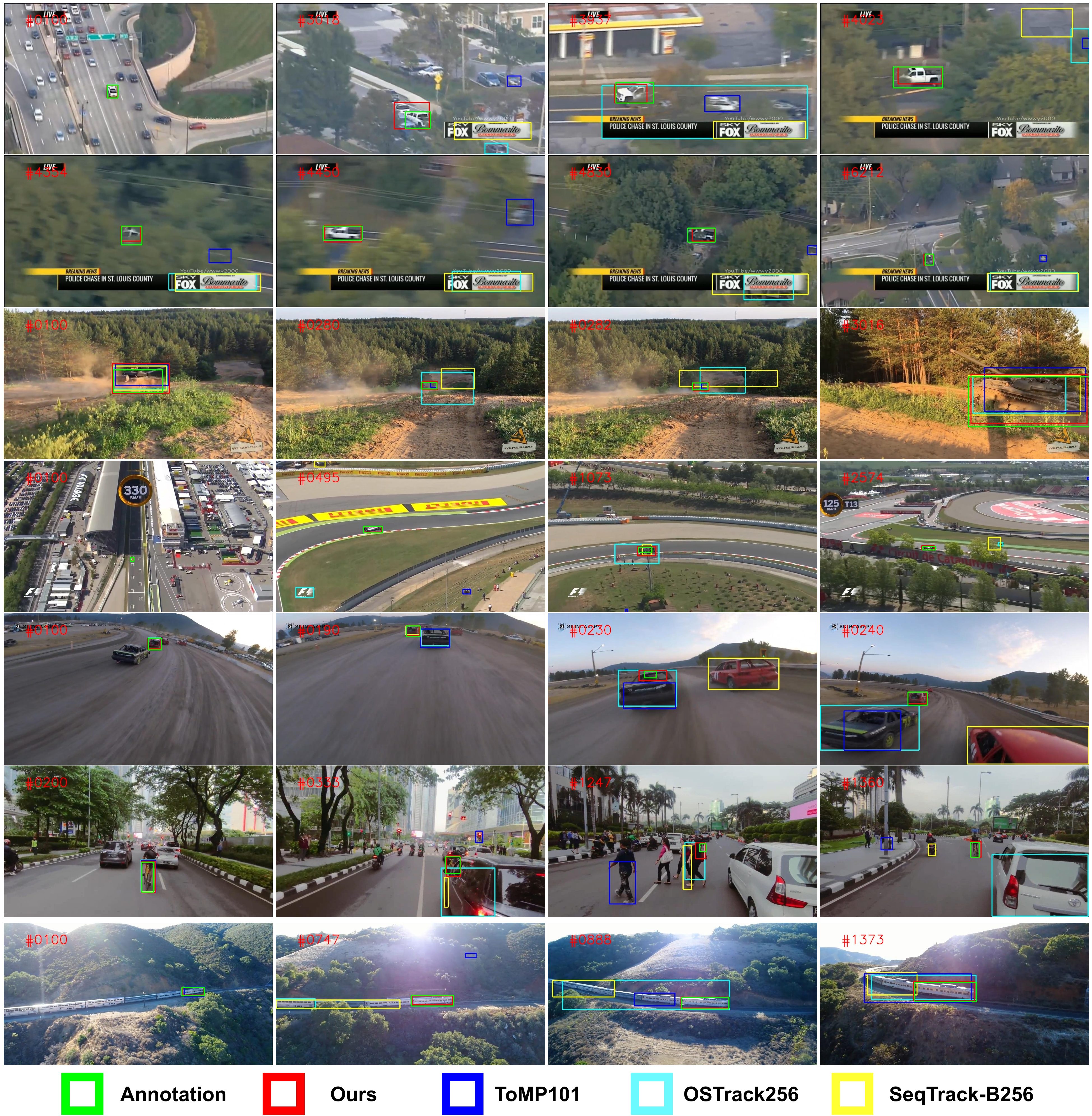}   \vspace{0mm}
\caption{ \textcolor{black}{Tracking results visualization of different trackers.}} \vspace{0mm}
\label{visualization}  
\end{figure}

\minisection{Tracking Visualization.}
As shown in~\Cref{visualization}, we provide a comparative analysis with current state-of-the-art single-branch trackers OSTrack256 and SeqTrack-B256, as well as the dual-branch tracker ToMP101. 
\textcolor{black}{In the intelligent vehicles field, there are many common challenges in real-world applications. The visualization results highlight that AViTMP exhibits superior performance, particularly in long-term tracking and complex scenes in autonomous driving. For instance, in rows 1 and 2, other methods struggle to track the truck after frame \#3018 due to challenging attributes like fast motion, heavy background occlusion, motion blur, and similar distractors, while our AViTMP achieves robust prediction results under these challenges. In row 3, when the tank undergoes a drastic scale change, conventional methods yield a rough bounding box, while AViTMP achieves a more precise and small-scale bounding box. Row 4 showcases a small vehicle fast-moving in a surveillance camera, where AViTMP demonstrates robust and refined tracking bounding boxes even when the target covers only a small region and a few pixels in each frame. In row 5, our method overcomes interference from similar red vehicles and still tracks the initial red one but not the others. In contrast, other methods wrongly track the other red vehicle, resulting in failed tracking in the following thousands of frames. Row 6 presents occlusion scenarios caused by similar pedestrians, common in vehicle driving recorders. These visualization results demonstrate the robustness and significant performance of AViTMP in the intelligent vehicle field and applications. 
}

\begin{figure} 
\centering   
\includegraphics[width=0.9\columnwidth]{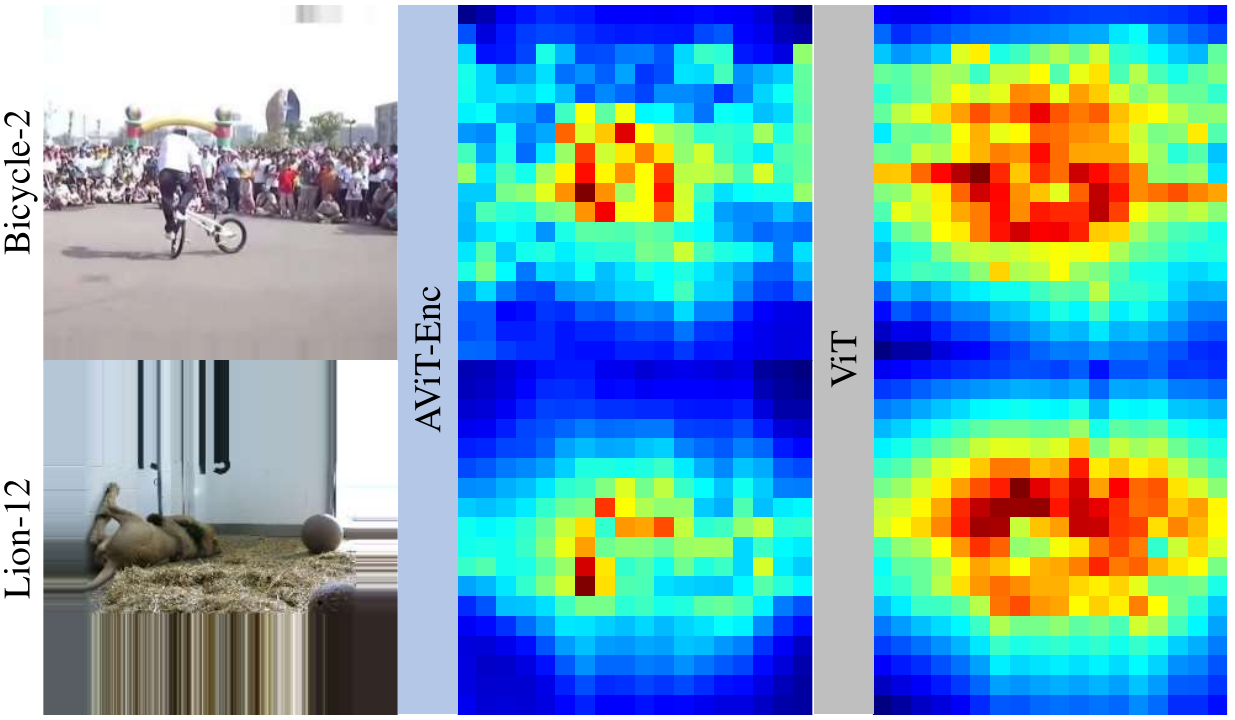}  \vspace{0mm}
\caption{  Visualizing features of test frames comparing AViT-Enc and standard ViT. The first column displays the input test frames, the second column shows the ViT-Enc features and the third column the ViT features. Features are aggregated along the channel dimension for visualization. The ViT-Enc features are more precisely localising the relevant objects. } 
\label{attn_vis} 
\end{figure}

\minisection{Feature Visualization}.
 To qualitatively compare the distribution of AViT-Enc features, a visualization of the output features from both vanilla ViT and AViT-Enc has been presented. 
As shown in ~\Cref{attn_vis}, we present a visualization of the output features from AViT-Enc components. During the testing phase, the test frame undergoes resizing and padding using nearby pixels. It is observed that the feature map of AViT exhibits a more precise and refined concentration on the target, while the ViT feature yields a more extensive representation and gets high responses in a variety of regions.

\minisection{\textcolor{black}{Attributes Analysis.}}
\textcolor{black}{For a more detailed comparison, ~\Cref{sup:tab:lasot_attributes} offers 12 attribute-specific performances compared to nine state-of-the-art methods. Our method performs well on most attributes. Especially, AViTMP performs much better on Illumination Variation (+1.5), Viewpoint Change (+1.9),  Full Occlusion (+1.5),  Fast Motion (+1.9), Out-of-View (+2.2) and Low Resolution (+1.9) compared with the second-best. Across all 14 attributes, AViTMP secures 9 first-best results and 5 second-best results, affirming its widespread efficacy. Through comprehensive analysis, we assert that AViTMP attains state-of-the-art performance across many attributes in the context of long-term tracking.
}

\input{tables/avit_ablation}

\subsection{Ablation and Analysis}
\minisection{Network Architecture.}
To analyze the effect of the tailored AViT-Enc encoder and TMP decoder, we train different variants of the encoder and decoder to ablate their roles. As shown in ~\Cref{tab:avit_ablation}, we report results for three variant encoders and two decoder parts.
As we can observe, the vanilla ViT (\#1) without Adaptor and joint state embedding sets the lowest performance in these encoder variants.
In \#2 row, as we employ the joint state embedding for each vanilla ViT layer, AUC improves 1.4\%/0.8\% in LaSOT and LaSOTExtSub, respectively.
While only embedding Adaptor module (\#3 row) into ViT, AUC outperforms the baseline with 3.4\%/2.7\%, showing the effectiveness and powerful ability of the Adaptor in contributing to the tracking-tailored backbone.
Finally, after combining joint state embedding and Adaptor (\#5 row) to build our AViT-Enc, we achieved the best performance compared with the vanilla ViT. 
\Cref{Framework_performance_compare} also proves the powerful advantages of our tailored AViT. 
\#4 shows the ablation of our dense-fusion decoder in TMP head. After replacing DF-Dec with the baseline decoder in DETR~\cite{carion2020detr}, AUC scores decrease around 0.3\%/0.4\% compared with our DF-Dec (\#5).
\textcolor{black}{
Additionally, as shown in ~\Cref{tab:inference} \#1 AViTMP achieves 68.0\% without any inference strategies. Current SOTA methods all use the template update strategy to improve performance. When removing the template update method, MixFormer and ToMP101 get a performance of 66.6\% and 65.7\%, still lower than ours without any inference strategy (68.0\%). 
The above comparison shows the strength of the proposed tracking-specific transformer network. }

\textcolor{black}{
In our decoder DF-Dec, the only hyperparameter is the layer number of ZSA+ZCA+FFN, we have added the ablation study to verify the performance. As shown in~\Cref{tab:layer_num}, with four layers, our method achieves the best performance with a fast inference speed. With six ZSA+ZCA+FFN layers, the speed will decrease by 10 FPS on the A40 GPU still being real-time, however it would no longer be real-time for weaker GPUs (RTX 2080Ti and 3090). Therefore, we use four layers to achieve the best balance between performance and running speed. 
}

\begin{table}[!t]
\centering
\caption{
\textcolor{black}{
Ablation on LaSOT and LaSOTExtSub over DF-Dec layer numbers. } } 
\setlength{\tabcolsep}{1mm}{
\begin{tabular}{c|llll}
\toprule
Layer Nums     & LaSOT    &LaSOTExtSub   & FPS      \\
\midrule
 2   & \multicolumn{1}{c}{69.8}   & \multicolumn{1}{c}{49.9}    & \multicolumn{1}{c}{\textbf{48}} \\   
 4     & \multicolumn{1}{c}{\textbf{70.7}}     & \multicolumn{1}{c}{50.2}   & \multicolumn{1}{c}{40}  \\   
 6    & \multicolumn{1}{c}{70.6}     & \multicolumn{1}{c}{\textbf{50.3}}   & \multicolumn{1}{c}{30}  \\   
\bottomrule
\end{tabular}
} 
\label{tab:layer_num}
\end{table}

\begin{figure}[t]
\centering  
\includegraphics[width=1\columnwidth]{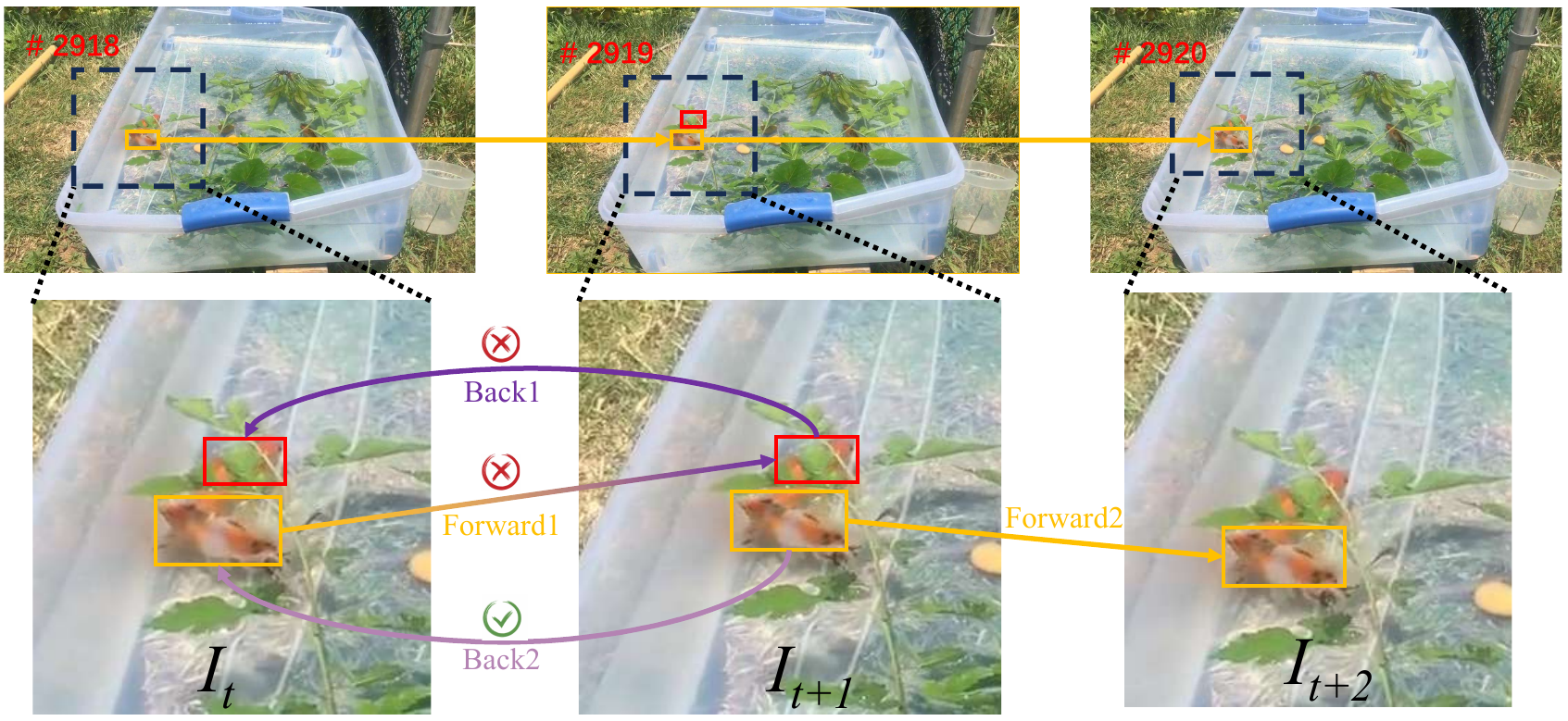} 
\caption{
Detailed process of CycleTrack inference strategy.
In frame \#2919, the forward tracking mistakenly predicts a distractor object (in \textcolor{red}{red} box), while CycleTrack corrects it to the right target (in \textcolor{orange}{orange} box).
}
\label{inference_vis} 
\end{figure}

\noindent
\textbf{Inference Strategies.}
During inference, we assess the effectiveness of our strategies in ~\Cref{tab:inference}. 
Upon the introduction of CycleTrack (\#2 \textit{vs.} \#1), AUC improves by 0.7\% on average with virtually no additional inference cost and speed influence (-1 FPS).
By updating the two training frames during inference (\#3  \textit{vs.} \#1), AUC improves by 2.1\%/1.5\% with only 2 FPS speed cost, which proves particularly effective in long-term tracking, especially with obviously scale-fluctuation, deformation, and poor-quality initial frame situations.
Consequently, we assert that the strategy of updating dual frames is beneficial for enhancing robustness in long-term tracking. 
Finally, with the amalgamation of these two strategies (\#4 \textit{vs.} \#1), AViTMP surpasses our base with 2.7\%/2.0\%, while incurring a minimal cost of 2 FPS. \textcolor{black}{Note that these strategies don't bring any network training cost.}
As shown in ~\Cref{inference_vis}, CycleTrack effectively rectifies erroneous predictions based on temporal cycle consistency under distractor scenarios, choosing the higher temporal consistency prediction result as the final location.

\input{tables/inference}

\minisection{Tracking Speed.} 
Our method achieves around 40 FPS with the parameter 217.9M on A40 GPU.
As shown in ~\Cref{tab:speed}, we run our method on different GPU types to check the speed for fair comparison. 
Using the same 2080Ti GPU, our method achieves significantly better performance with a bit slower speed compared with discriminate methods ToMP101 and KeepTrack. 
With running on RTX3090 device, our method is a bit slower than GRM (38FPS \textit{vs.} 45FPS) with outperforming performance.

\input{tables/speed}

\section{Conclusion} 
In this paper, we introduce a novel method called AViTMP, which operates as an \textcolor{black}{specific-designed} adaptive vision transformer model predictor for single-branch visual tracking. 
\textcolor{black}{We propose the first tracking-tailored AViT backbone to solve the lack of image-related inductive biases in vanilla ViT. }
By \textcolor{black}{adding} a joint state embedding, AViTMP encodes target features with a location-prior. Next, the \textcolor{black}{ transformer model predictor} estimates model weights to predict object locations in test frames.
With the seamless integration of the AViT encoder and discriminative model predictor, our approach harmoniously merges the strengths of single-branch trackers with those of discriminative models, establishing a cutting-edge paradigm in visual tracking.
Furthermore, with our proposed CycleTrack strategy, we also refine the inference process to ensure the integration of temporal consistency and robustness inference within sequences.
Comprehensive experiments and analyses validate the effectiveness of our proposed method.

\minisection{\textcolor{black}{Future Work.}} \textcolor{black}{
By merging discriminate trackers with single-branch methods, our approach incorporates the strengths and weaknesses of each pipeline to optimize performance. 
In our future work, we focus on the efficiency of the method, aiming to reduce memory usage, training and inference time. This would also allow extending training to use larger backbone~\cite{mixformer,gao2023GRM} and higher resolution~\cite{chen2023seqtrack,ye2022OSTrack}, and thereby further improve performance. Furthermore, we are interested in extending our method to multi-object tracking, especially our inference strategy. 
}

\section{Acknowledgments}
\noindent 
We acknowledge the support from the Spanish Government funding for projects PID2022-143257NB-I00, TED2021-132513B-I00 funded by MCIN/AEI/10.13039/501100011033 and by FSE+ and the European Union NextGenerationEU/PRTR, and the CERCA Programme of Generalitat de Catalunya.
This work is also jointly supported by Frontier Research Fund of Institute of Optics and Electronics, China Academy of Sciences (Grant No. C21K005) and National Natural Science Foundation of China(Grant No.62101529).   
Chuanming further acknowledges Chinese Scholarship Council (CSC) No.202204910331.

\bibliographystyle{IEEEtran}
\bibliography{refs}

\begin{IEEEbiography}[{\includegraphics[width=1in,height=1.25in,clip,keepaspectratio]{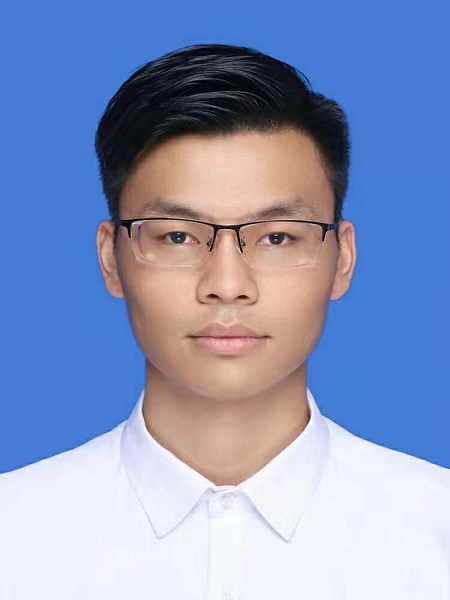}}]{Chuanming Tang}
received the B.S degree in electronic and information engineering from Southwest University, China, in 2019. He is currently pursuing the PhD degree with the University of Chinese Academy of Sciences, Beijing, China and the Institute of Optics and Electronics, Chinese Academy of Sciences, Chengdu, China. 
He is currently a visiting PhD student in Computer Vision Center, Barcelona, Spain.
His current research interests include visual tracking, vision transformers and image generation.
\end{IEEEbiography}

\begin{IEEEbiography}[{\includegraphics[width=1in,height=1.25in,clip,keepaspectratio]{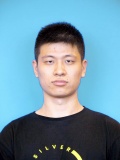}}]{Kai Wang} is a postdoctoral researcher at Computer Vision Center, UAB. Before he obtained the Ph.D. degree from Computer Vision Center, UAB in 2022 under the supervision of Joost van de Weijer. He received the master degree in image processing from Jilin University in 2017 and the bachelor degree from Jilin University in 2014. His main research interests include continual learning, knowledge distillation, domain adaptation and vision transformers.
\end{IEEEbiography}

\begin{IEEEbiography}[{\includegraphics[width=1in,height=1.25in,clip,keepaspectratio]{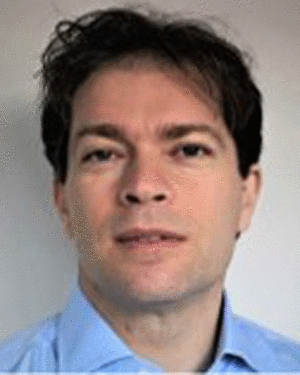}}]{Joost van de Weijer} received the PhD degree from the University of Amsterdam, Amsterdam, Netherlands, in 2005. He was a Marie Curie Intra-European fellow with INRIA Rhone-Alpes, France, and from 2008 to 2012, he was a Ramon y Cajal fellow with the Universitat Autònoma de Barcelona, Barcelona, Spain, where he is currently a senior scientist with the Computer Vision Center and leader of the Learning and Machine Perception (LAMP) Team. His main research directions are color in computer vision, continual learning, active learning, and domain adaptation.
\end{IEEEbiography}

\begin{IEEEbiography}[{\includegraphics[width=1in,height=1.25in,clip,keepaspectratio]{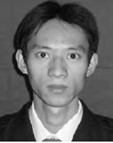}}]{Jianlin Zhang}
 received the Ph.D. degree in signal and information processing from the University of Chinese Academy of Sciences, in 2008. He is currently a Full Professor with the Institute of Optics and Electronics, Chinese Academy of Sciences, Chengdu, China. His research interests include object detection and tracking, computer vision, machine learning, and artificial intelligence. He has published more than 20 papers, conference papers in those areas.
\end{IEEEbiography}

\begin{IEEEbiography}[{\includegraphics[width=1in,height=1.25in,clip,keepaspectratio]{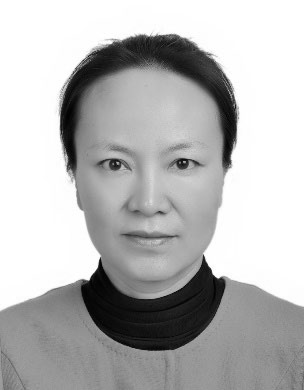}}]{Yongmei Huang}
 received the B.S. degree from the Department of Automation, University of Electronic Science and Technology of China, in 1989, and the Ph.D. degree from the Institute of Optics and Electronics, Chinese Academy of Sciences, in 2005. Since 2005, she has been a Professor with the University of Chinese Academy of Sciences. She has published more than 60 papers in refereed conferences and journals. Her research interests include quantum teleportation, laser communication, object detection, and tracking. She received the Distinguished Scientific Achievement Award twice from the Chinese Academy of Sciences in 2011 and 2019. 
\end{IEEEbiography}

\vfill

\end{document}

%% file: tables/lasot.tex
\begin{table*}[!t]
	\centering
 	\caption{Comparison on the LaSOT~\cite{lasot} test set ordered by AUC. Only trackers of similar complexity are included.
	}  
	\newcommand{\best}[1]{\textbf{\textcolor{red}{#1}}}
	\newcommand{\scnd}[1]{\textbf{\textcolor{blue}{#1}}}
	\newcommand{\opt}[1]{\textbf{\textcolor{violet}{#1}}}
	\newcommand{\fast}[1]{\textbf{\textcolor{orange}{#1}}}
	\newcommand{\dist}{\hspace{3pt}}%
	\resizebox{1.00\linewidth}{!}{%
        \begin{tabular}{l@{\dist}c@{\dist}c@{\dist}c@{\dist}c@{\dist}c@{\dist}c@{\dist}c@{\dist}c@{\dist}c@{\dist}c@{\dist}c@{\dist}c@{\dist}c@{\dist}c@{\dist}c@{\dist}c@{\dist}c@{\dist}c@{\dist}c@{\dist}c@{\dist}c@{\dist}c@{\dist}c@{\dist}c@{\dist}c@{\dist}c@{\dist}c@{\dist}c@{\dist}c@{\dist}}
        	\toprule
        	           & \textbf{Ours}  &SeqTrack &  &Sim  &MixFormer &OSTrack  & ToMP  & ToMP  & STARK & Keep  & STARK & Alpha &        & Siam  & Tr   & Super   & Pr    \\
        	           &  &B256 &GRM &-B/16  &22k &256 & 101  & 50  & ST101 & Track & ST50  &Refine & TransT & R-CNN & DiMP & DiMP  & DiMP  \\
        	           &     &\cite{chen2023seqtrack}   &\cite{gao2023GRM}  &\cite{chen2022SimTrack}  &\cite{mixformer}  &\cite{ye2022OSTrack}  & \cite{tomp}      & \cite{tomp}               &\cite{stark} & \cite{keeptrack} &\cite{stark} & \cite{Yan_2021_CVPR_AlphaRefine} & \cite{chen2021transt} & \cite{siamr-cnn} & \cite{trdimp} & \cite{Danelljan_2019_github_pytracking}  & \cite{prdimp}  \\
        	\midrule
        	Success (AUC)  & \best{70.7}  &\scnd{69.9} &\scnd{69.9}  &69.3  &{69.2}&69.1   &68.5  & {67.6} & 67.1        & 67.1 & 66.4 & 65.3 & 64.9 & 64.8 & 63.9 & 63.1   & 59.8 \\        	
        	Norm. Prec     & \best{80.5}  &\scnd{79.7} &79.3 &78.5  &78.7  &78.7 &78.7 & {78.0} & 76.9        & 77.2 & 76.3 & 73.2 & 73.8 & 72.2 & 73.0 & 72.2   & 68.8 \\ 
         Precision      & \scnd{75.9} &\best{76.3} &75.8  &75.2 &74.7  &- &73.5  & {72.2} & {72.2} & 70.2 & 71.2 & 68.0 & 69.0 & 68.4 & 66.3 & 65.3    & 60.8 \\
\bottomrule
        \end{tabular}
	}  
	\label{tab:lasot}%
\end{table*}

%% file: tables/lasotext.tex
\begin{table*}[!t]
	\centering  
	\caption{Comparison on the LaSOTExtSub~\cite{Fan_2020_IJCV_Lasot_ext} test set ordered by AUC. Only trackers of similar complexity are included.
	}   
	\newcommand{\best}[1]{\textbf{\textcolor{red}{#1}}}
	\newcommand{\scnd}[1]{\textbf{\textcolor{blue}{#1}}}
	\newcommand{\opt}[1]{\textbf{\textcolor{violet}{#1}}}
	\newcommand{\fast}[1]{\textbf{\textcolor{orange}{#1}}}
	\newcommand{\dist}{\hspace{3pt}}%
	\resizebox{1.00\linewidth}{!}{%
        \begin{tabular}{l@{\dist}c@{\dist}c@{\dist}c@{\dist}c@{\dist}c@{\dist}c@{\dist}c@{\dist}c@{\dist}c@{\dist}c@{\dist}c@{\dist}c@{\dist}c@{\dist}c@{\dist}c@{\dist}c@{\dist}c@{\dist}c@{\dist}c@{\dist}c@{\dist}c@{\dist}c@{\dist}c@{\dist}c@{\dist}c@{\dist}c@{\dist}c@{\dist}c@{\dist}c@{\dist}}
        	\toprule
        	           & \textbf{Ours} &SeqTrack &SwinTrack &Keep  &OSTrack &AiA & ToMP  & ToMP   & LTMU &   & SiamRPN   &     & Auto          \\
        	           &               &B256       & B384       &Track   &256  &Track  & 101  & 50    & DiMP     &DiMP    & ++   &ATOM   &Match   \\
        	           &     &\cite{chen2023seqtrack}   &\cite{swintrack}  & \cite{keeptrack}  &\cite{ye2022OSTrack}      & \cite{tomp}   &\cite{gao2022aiatrack}    &\cite{tomp}  & \cite{ltmu} & \cite{dimp} & \cite{li2019siamrpnpp} & \cite{atom} & \cite{automatch}   \\
        	\midrule
        	Success (AUC)  & \best{50.2}  &\scnd{49.5} &49.1 &48.2 &47.4  &46.8 &45.9  &45.4    & 41.4 & 39.2 & 34.0 & 37.6 & 37.6 \\        	
        	Norm. Prec     & \best{62.6}  &\scnd{60.8}  &- &61.7 &{57.3} & 54.4  &58.1 &57.6      & 49.9   & 47.6 & 41.6 & 45.9 & -  \\
         Precision      & \best{57.7} &\scnd{56.3}  & 55.6 & 54.5 &53.3  & 54.2 &52.7 
 &51.9     &47.3  & 45.1 & 39.6 & 43.0 & 43.0 \\
\bottomrule
        \end{tabular}
	}
	\label{tab:lasotext}%
\end{table*}

%% file: tables/avist.tex
\begin{table*}[!t]
	\centering
 	\caption{Comparison on the AVisT~\cite{noman2022avist} test set ordered by AUC.
	} 
	\newcommand{\best}[1]{\textbf{\textcolor{red}{#1}}}
	\newcommand{\scnd}[1]{\textbf{\textcolor{blue}{#1}}}
	\newcommand{\opt}[1]{\textbf{\textcolor{violet}{#1}}}
	\newcommand{\fast}[1]{\textbf{\textcolor{orange}{#1}}}
	\newcommand{\dist}{\hspace{3pt}}%
	\resizebox{1.00\linewidth}{!}{%
        \begin{tabular}{l@{\dist}c@{\dist}c@{\dist}c@{\dist}c@{\dist}c@{\dist}c@{\dist}c@{\dist}c@{\dist}c@{\dist}c@{\dist}c@{\dist}c@{\dist}c@{\dist}c@{\dist}c@{\dist}c@{\dist}c@{\dist}c@{\dist}c@{\dist}c@{\dist}c@{\dist}c@{\dist}c@{\dist}c@{\dist}c@{\dist}c@{\dist}c@{\dist}c@{\dist}c@{\dist}}
        	\toprule
        	           &  &  &MixFormer   &ToMP  &STARK &ToMP   &     &Keep    &Alpha    &Tr  &Pr &  & & \\
                          &\textbf{Ours}  &GRM  &22k  &50  &ST50  &101   &RTS     &Track    &Refine    &DiMP  &DiMP &DiMP &Ocean &ATOM \\
        	           &  &\cite{gao2023GRM}  &\cite{mixformer} &\cite{tomp} &\cite{stark}  &\cite{tomp} &\cite{paul2022rts} &\cite{keeptrack} &\cite{Yan_2021_CVPR_AlphaRefine}    &\cite{trdimp}  &\cite{prdimp} &\cite{dimp}   &\cite{zhang2020ocean} &\cite{atom}    \\
        	\midrule
        	Success (AUC)  & \best{54.9} &\scnd{54.5}  &53.7 &51.6 &51.1 &50.9  &50.8   &49.4     &49.6   &48.1  &43.3  &41.9  &38.9 &38.6  \\        	
        	OP50     & \best{64.0}       &\scnd{63.1}  &63.0 &59.5 &59.2 &58.8  &55.7   &56.3     &55.7   &55.3  &48.0  &45.7  &43.6 &41.5  \\
\bottomrule
        \end{tabular}
	} 
	\label{tab:avist}%
\end{table*}

%% file: tables/vot2020.tex
\begin{table}[!t]
	\centering  
	\caption{Comparison to bounding box only methods on the VOT2020~\cite{Kristan_2020_ECCVW_VOT2020} dataset in terms of EAO score.	}  
	\newcommand{\best}[1]{\textbf{\textcolor{red}{#1}}}
	\newcommand{\scnd}[1]{\textbf{\textcolor{blue}{#1}}}
	\newcommand{\dist}{\hspace{5pt}}%
	\resizebox{1.00\linewidth}{!}{%
        \begin{tabular}{l@{\dist}c@{\dist}c@{\dist}c@{\dist}c@{\dist}c@{\dist}c@{\dist}c@{\dist}c@{\dist}c@{\dist}c@{\dist}c@{\dist}c@{\dist}c@{\dist}c@{\dist}c@{\dist}c@{\dist}c@{\dist}c@{\dist}c@{\dist}c@{\dist}c@{\dist}c@{\dist}}
        	\toprule
        	    & \textbf{Ours}     &SeqTrack & ToMP & STARK        & Super & CSWin & STARK   &ToMP    & Tr    \\
        	     &    &B256  & {101}    & ST50         & DiMP  &TT  & ST101     & {50}  &DiMP \\
        	    &    &\cite{chen2023seqtrack}   &\cite{tomp} & \cite{stark} & \cite{Danelljan_2019_github_pytracking,Kristan_2020_ECCVW_VOT2020}  &\cite{cswintt}  & \cite{stark} &\cite{tomp}  &\cite{trdimp} \\          
        	\midrule
        	EAO   &\best{0.314}   &\scnd{0.312}   & {0.309}       & {0.308}  & 0.305 &0.304 & 0.303        & 0.303  & 0.300   \\
         Accuracy &0.446  & 0.473 & 0.453              & 0.478.        & 0.477 & \scnd{0.480} & \best{0.481}   & 0.453  & 0.471  \\
        	Robustness  & \best{ 0.840}   & 0.806   & \scnd{0.814}         & {0.799}  & 0.728  & 0.787  & 0.775      & 0.789  & 0.782  \\
         \bottomrule

        \end{tabular}
	}  
	\label{tab:vot2020}%
\end{table}

%% file: tables/uav_tnl2k.tex
\begin{table}[!t]
	\centering
	\caption{Comparison with the state-of-the-art methods on UAV123~\cite{Mueller_2016_ECCV_UAV123} and TNL2k~\cite{tnl2k}  in terms of AUC.
	} 
	\newcommand{\best}[1]{\textbf{\textcolor{red}{#1}}}
	\newcommand{\scnd}[1]{\textbf{\textcolor{blue}{#1}}}
	\newcommand{\dist}{\hspace{5pt}}%
	\resizebox{1.00\linewidth}{!}{%
        \begin{tabular}{l@{\dist}c@{\dist}c@{\dist}c@{\dist}c@{\dist}c@{\dist}c@{\dist}c@{\dist}c@{\dist}c@{\dist}c@{\dist}c@{\dist}c@{\dist}c@{\dist}c@{\dist}c@{\dist}c@{\dist}c@{\dist}c@{\dist}c@{\dist}c@{\dist}c@{\dist}c@{\dist}c@{\dist}c@{\dist}c@{\dist}c@{\dist}c@{\dist}c@{\dist}c@{\dist}}
        	\toprule
        	    & \textbf{}     &SeqTrack  & Keep   & OSTrack  &   & Super  & Pr    & STM     & Siam  &      &      \\
        	     & \textbf{Ours}        &B256 &Track   &256    &TransT   & DiMP   & DiMP  & Track      & R-CNN & DiMP  \\
        	    &  &\cite{chen2023seqtrack}  &\cite{keeptrack}   &\cite{ye2022OSTrack}  &\cite{chen2021transt}    & \cite{Danelljan_2019_github_pytracking} & \cite{prdimp} & \cite{fu2021stmtrack}  & \cite{siamr-cnn}  & \cite{dimp}  \\          
        	\midrule
        	UAV123  &\best{70.1}  &69.2 &\scnd{69.7}  &68.3       &69.1  & 67.7   & 68.0 & 64.7        & 64.9   & 65.3 \\
                TNL2k   &\scnd{54.5}    &\best{54.9}  &-    &54.3  &50.7     &49.2       &47.0     &38.4            &52.3         & 44.7           \\ 
         \bottomrule

        \end{tabular}
	}  
	\label{tab:uav_tnl}%
\end{table}

%% file: tables/lasot_attributes.tex
\begin{table*}[t]
	\centering
	\caption{ Comparison of different attributes analysis  with the state-of-the-art on LaSOT~\cite{lasot} benchmark.   }
	\newcommand{\best}[1]{\textbf{\textcolor{red}{#1}}}
	\newcommand{\scnd}[1]{\textbf{\textcolor{blue}{#1}}}
	\newcommand{\dist}{\hspace{4pt}}%
	\resizebox{1.00\textwidth}{!}{%
        \begin{tabular}{l@{\dist}c@{\dist}c@{\dist}c@{\dist}c@{\dist}c@{\dist}c@{\dist}c@{\dist}c@{\dist}c@{\dist}c@{\dist}c@{\dist}c@{\dist}c@{\dist}c@{\dist}|c@{\dist}}
        	\toprule
        	                       & Illumination & Partial     &                & Motion         & Camera      &             & Background & Viewpoint   & Scale        & Full        & Fast        &             & Low         & Aspect        &       \\
        	                       & Variation    & Occlusion   & Deformation    & Blur           & Motion      & Rotation    & Clutter    & Change      & Variation    & Occlusion   & Motion      & Out-of-View & Resolution  & Ration Change & Total \\
 
        	\midrule
            TransT                 & 65.2         & 62.0        & 67.0           & 63.0           & 67.2        & 64.3        & 57.9        & 61.7        & 64.6        & 55.3        & 51.0        & 58.2        & 56.4        & 63.2          & 64.9 \\
            STARK-ST101            & 67.5         & {65.1} & 68.3           & 64.5           & 69.5        & 66.6        & 57.4        & {68.8} & 66.8        & 58.9        & 54.2        & 63.3        & 59.6        & 65.6          & 67.1 \\
            KeepTrack              & {69.7}  & 64.1        & 67.0           & {66.7}    & {71.0} & 65.3        & {61.2} & 66.9        & 66.8        & {60.1} & {57.7} & {64.1} & {62.0} & 65.9          & 67.1 \\
            ToMP-50       & 66.8         & 64.9        & {68.5}    & 64.6           & 70.2        & {67.3} & 59.1        & 67.2        & {67.5} & {59.3} & 56.1        & {63.7} & 61.1        & {66.5}   & {67.6} \\
            ToMP-101      & 69.0         & {65.3} & {69.4}    & {65.2}    & {71.7} & {67.8} & {61.5} & {69.2} & {68.4} & 59.1        & \scnd{57.9} & {64.1} & {62.5} & {67.2}   & {68.5} \\
            OSTrack256        & 68.7      & 66.6      & 71.2      & 66.4      & 72.0      & 68.6      & 61.5      & 69.1      & 69.0      & 59.5      & 55.7      & 63.2      & 61.7      & 67.4      &  69.1  \\
            MixFormer22k      & 69.6      & 66.5      & 69.7      & 66.5      & 71.6      & 68.6      & 59.9      & 70.6      & 68.9      & 61.4      & 56.6      & 64.4      & \scnd{62.9}      & 67.7      &  69.2 \\
            GRM            & \scnd{70.0}      & \best{67.9}      & \best{72.3}      & 66.7      & 71.9      & 69.3      & 62.1      & 68.4      & \scnd{69.8}      & 60.8      & 55.3      & 64.5      & 62.5      & 68.2      & \scnd{69.9}  \\
            SeqTrack-B256          & 68.6      & 67.7      & 70.8      & \best{69.2}      & \best{73.3}      & \scnd{69.8}      & \best{62.9}      & \scnd{71.3}      & 69.6      & \scnd{61.6}      & 57.8      & \scnd{64.6}      & 62.2      & \scnd{68.3}      & \scnd{69.9}  \\
            \textbf{Ours}            & \best{71.5}      & \scnd{67.8}      & \scnd{71.7}      & \scnd{66.8}      & \scnd{73.2}      & \best{70.4}      & \scnd{62.8}      & \best{73.2}      & \best{70.3}      & \best{63.3}      & \best{59.8}      & \best{66.8}      & \best{64.8}      & \best{69.3}      & \best{70.7}  \\
            \bottomrule
        \end{tabular}
	}  
	\label{sup:tab:lasot_attributes}
\end{table*}

%% file: tables/trackingnet.tex
\begin{table}[t]
	\centering
         \vspace{-2mm}
	\caption{Comparison with other method on TrackingNet~\cite{trackingnet}.
	} 
	\newcommand{\best}[1]{\textbf{\textcolor{red}{#1}}}
	\newcommand{\scnd}[1]{\textbf{\textcolor{blue}{#1}}}
	\newcommand{\dist}{\hspace{3pt}}%
	\resizebox{1.00\linewidth}{!}{%
        \begin{tabular}{l@{\dist}c@{\dist}c@{\dist}c@{\dist}c@{\dist}c@{\dist}c@{\dist}c@{\dist}c@{\dist}c@{\dist}c@{\dist}c@{\dist}c@{\dist}c@{\dist}c@{\dist}c@{\dist}c@{\dist}c@{\dist}c@{\dist}c@{\dist}c@{\dist}c@{\dist}c@{\dist}c@{\dist}c@{\dist}c@{\dist}c@{\dist}c@{\dist}c@{\dist}c@{\dist}}
        	\toprule
        	      &  &SeqTrack  & {ToMP}& STARK &        & Siam  & Alpha      & Tr   & Keep  & Pr  \\
        	   &\textbf{Ours}  &B256 & {101}   & ST101 & TransT  & R-CNN &Refine   & DiMP & Track  & DiMP  \\
        	      &  &\cite{chen2023seqtrack}   &\cite{tomp}                   & \cite{stark} &\cite{chen2021transt} & \cite{siamr-cnn} & \cite{Yan_2021_CVPR_AlphaRefine}   & \cite{trdimp} & \cite{keeptrack} & \cite{prdimp} \\
        	\midrule
        	
        	PRE    &\scnd{80.7}  &\best{ 82.2}    & 78.9    & -           & {80.3}    & 80.0 & 78.3   & 73.1 & 73.8 & 70.4 \\
        	NPR    &\scnd{87.1}  &\best{88.3}   & 86.4    & {86.9} & {86.7}  & 85.4 & 85.6   & 83.3 & 83.5 & 81.6 \\
        	AUC   &\scnd{82.8}  & \best{83.3}   &{81.5} & {82.0} & 81.4     & 81.2 & 80.5  & 78.4 & 78.1 & 75.8  \\\bottomrule
        \end{tabular}
	} 
	\label{trackingnettable}%
\end{table}


%% file: tables/vot2020_mask.tex
\begin{table}[t]
	\centering
	\vspace{-2mm}
	\caption{Comparison to segmentation only methods on VOT2020~\cite{Kristan_2020_ECCVW_VOT2020}.
	}
	\newcommand{\best}[1]{\textbf{\textcolor{red}{#1}}}
	\newcommand{\scnd}[1]{\textbf{\textcolor{blue}{#1}}}
	\newcommand{\dist}{\hspace{5pt}}%
	\resizebox{1.00\linewidth}{!}{%
        \begin{tabular}{l@{\dist}c@{\dist}c@{\dist}c@{\dist}c@{\dist}c@{\dist}c@{\dist}c@{\dist}c@{\dist}c@{\dist}c@{\dist}c@{\dist}c@{\dist}c@{\dist}c@{\dist}c@{\dist}c@{\dist}c@{\dist}c@{\dist}c@{\dist}c@{\dist}c@{\dist}c@{\dist}c@{\dist}c@{\dist}c@{\dist}c@{\dist}c@{\dist}c@{\dist}c@{\dist}}
        	\toprule
        	       &\textbf{Ours}    & ToMP   &ToMP    &SeqTrack     & STARK        & STARK      & Alpha  &      &       \\
        	       &\textbf{+HQ-SAM}    & {101+AR} & {50 +AR} &B256+AR  & ST50+AR      & ST101+AR   & Refine & AFOD   \\
        	       &    &\cite{tomp} &\cite{tomp} & \cite{chen2023seqtrack} & \cite{stark} & \cite{stark}  & \cite{Yan_2021_CVPR_AlphaRefine} & \cite{Kristan_2020_ECCVW_VOT2020}    \\          
        	\midrule
        	EAO    &0.504    & 0.497 & 0.496 & \best{0.520} & \scnd{0.505}        & 0.497           & 0.482 & 0.472 \\
        	A &0.725  & 0.750 & 0.754 & -        & \scnd{0.759} & \best{0.763}    & 0.754 & 0.713    \\
        	R &\best{0.821}    & 0.798 & 0.793 & - & \scnd{0.817}        & 0.789        & 0.777 & 0.795   \\ 
            \bottomrule
        \end{tabular}
	}
	\label{tab:vot2020mask}%
\end{table}

%% file: tables/avit_ablation.tex
\begin{table}[t]
	\centering
 	\caption{Analysis of architecture component variants.  
B-Dec denotes the plain baseline decoder~\cite{carion2020detr}. $\Delta$ denotes the AUC change compared with the baseline (1st row). 
	}  
	\newcommand{\best}[1]{\textbf{\textcolor{red}{#1}}}
	\newcommand{\scnd}[1]{\textbf{\textcolor{blue}{#1}}}
	\newcommand{\dist}{\hspace{14pt}}%
	\newcommand{\yes}{\textcolor{black}{\checkmark}}
          \newcommand{\no}{\ding{55}}%
	   \resizebox{\columnwidth}{!}{%
        \begin{tabular} 
{l@{\dist\extracolsep{1pt}}|c@{\dist\extracolsep{1pt}}c@{\dist\extracolsep{1pt}}|c@{\dist\extracolsep{1pt}}c@{\dist\extracolsep{1pt}}|c@{\dist\extracolsep{1pt}}c@{\dist\extracolsep{1pt}}c@{\dist\extracolsep{1pt}}c@{\dist\extracolsep{1pt}}}
        	\toprule
        	        \#  &JSE  &Adaptor  & DF-Dec & B-Dec  & LaSOT &$\Delta_1$ & LaSOTExtSub    &$\Delta_2$ \\
        	\midrule
                 1  &     &    & \yes    &      &65.7 &-  &45.8 &- \\ 
                 2  & \yes   &   & \yes  &      & 67.1 &+1.4 &  46.6  &+0.8\\ 
                 3  &    & \yes  & \yes  &      &69.1   &+3.4   &48.5   &+2.7    \\ 
                 4  & \yes   & \yes  &  & \yes  &70.4 &+4.7 &47.8   &+2.0 \\ 
                 5  & \yes   & \yes  & \yes  &  & \textbf{70.7}  &+5.0 & \textbf{50.2}   &+4.4 \\ 
        \bottomrule
        \end{tabular}
	} 
	\label{tab:avit_ablation}%
\end{table}


%% file: tables/inference.tex
\begin{table}[t]
	\centering
	\caption{Ablation on LaSOT and LaSOTExtSub over inference strategies. $\Delta$  denotes the AUC change compared with the baseline (1st row). FPS measures the inference speed. DFU means Dual-Frame Update.
	}
	\newcommand{\best}[1]{\textbf{\textcolor{red}{#1}}}
	\newcommand{\scnd}[1]{\textbf{\textcolor{blue}{#1}}}
	\newcommand{\dist}{\hspace{14pt}}%
	\newcommand{\yes}{\textcolor{black}{\checkmark}}
         \newcommand{\no}{\ding{55}}%
	\resizebox{1.\columnwidth}{!}{%
        \begin{tabular}{c@{\dist}|c@{\dist}c@{\dist}|c@{\dist}c@{\dist}c@{\dist}c@{\dist}c@{\dist}}
        	\toprule
        	        \#  & CycleTrack  &DFU    & LaSOT  &$\Delta_1$ &LaSOTExtSub   &$\Delta_2$  & FPS \\
        	\midrule 
 1  &            &                         &68.0    & -         &48.2      & -      & 42  \\
 2  &\yes  &                         & 68.9        & +0.9       &48.7  &+0.5      & 41   \\
3  &   &\yes       &70.1    & +2.1     &49.7       & +1.5       & 40   \\
 4  &\yes   &\yes         &\textbf{70.7}     & +2.7    &\textbf{50.2}   & +2.0     & 40  \\
        \bottomrule
        \end{tabular}
	}
	\label{tab:inference}%
\end{table}

%% file: tables/speed.tex
\begin{table}[!t]
\centering
\caption{Tracking speed and performance comparision of different tracking pipelines. We have reported the numbers and GPU type provided by the authors. AUC is evaluated on LaSOT.
} 
\resizebox{\columnwidth}{!}{
\setlength{\tabcolsep}{1mm}{
\begin{tabular}{l|llllllll}
\toprule
    & \multicolumn{1}{c}{KeepTrack}  & \multicolumn{1}{c}{ToMP101}    & \multicolumn{1}{c}{MixFormer-22k}    & \multicolumn{1}{c}{GRM}    & \multicolumn{1}{c}{\textbf{Ours$_1$}}  & \multicolumn{1}{c}{\textbf{Ours$_2$}}  & \multicolumn{1}{c}{\textbf{Ours$_3$}}  \\
\midrule
FPS   & \multicolumn{1}{c}{18} & \multicolumn{1}{c}{20} & \multicolumn{1}{c}{25} &\multicolumn{1}{c}{45}   & \multicolumn{1}{c}{15} & \multicolumn{1}{c}{38}    & \multicolumn{1}{c}{40} \\  
GPU    & \multicolumn{1}{c}{2080Ti} & \multicolumn{1}{c}{2080Ti}  
 &\multicolumn{1}{c}{1080Ti} &\multicolumn{1}{c}{RTX3090}   & \multicolumn{1}{c}{2080Ti}  &\multicolumn{1}{c}{RTX3090} & \multicolumn{1}{c}{A40} \\ 
AUC    & \multicolumn{1}{c}{67.1} & \multicolumn{1}{c}{67.6}  &\multicolumn{1}{c}{70.1} &\multicolumn{1}{c}{69.9} & \multicolumn{1}{c}{70.7} & \multicolumn{1}{c}{70.6}  & \multicolumn{1}{c}{70.7}  \\    
\bottomrule
\end{tabular}
} }
\label{tab:speed}
\end{table}

%% file: Main.bbl
\begin{thebibliography}{10}
\providecommand{\url}[1]{#1}
\csname url@samestyle\endcsname
\providecommand{\newblock}{\relax}
\providecommand{\bibinfo}[2]{#2}
\providecommand{\BIBentrySTDinterwordspacing}{\spaceskip=0pt\relax}
\providecommand{\BIBentryALTinterwordstretchfactor}{4}
\providecommand{\BIBentryALTinterwordspacing}{\spaceskip=\fontdimen2\font plus
\BIBentryALTinterwordstretchfactor\fontdimen3\font minus \fontdimen4\font\relax}
\providecommand{\BIBforeignlanguage}[2]{{%
\expandafter\ifx\csname l@#1\endcsname\relax
\typeout{** WARNING: IEEEtran.bst: No hyphenation pattern has been}%
\typeout{** loaded for the language `#1'. Using the pattern for}%
\typeout{** the default language instead.}%
\else
\language=\csname l@#1\endcsname
\fi
#2}}
\providecommand{\BIBdecl}{\relax}
\BIBdecl

\bibitem{gunduz2019efficient}
G.~G{\"u}nd{\"u}z and T.~Acarman, ``Efficient multi-object tracking by strong associations on temporal window,'' \emph{IEEE Transactions on Intelligent Vehicles}, vol.~4, no.~3, pp. 447--455, 2019.

\bibitem{rangesh2019no}
A.~Rangesh and M.~M. Trivedi, ``No blind spots: Full-surround multi-object tracking for autonomous vehicles using cameras and lidars,'' \emph{IEEE Transactions on Intelligent Vehicles}, vol.~4, no.~4, pp. 588--599, 2019.

\bibitem{tang2023transformer}
C.~Tang, Q.~Hu, G.~Zhou, J.~Yao, J.~Zhang, Y.~Huang, and Q.~Ye, ``Transformer sub-patch matching for high-performance visual object tracking,'' \emph{IEEE Transactions on Intelligent Transportation Systems}, 2023.

\bibitem{javed2022visual}
S.~Javed, M.~Danelljan, F.~S. Khan, M.~H. Khan, M.~Felsberg, and J.~Matas, ``Visual object tracking with discriminative filters and siamese networks: a survey and outlook,'' \emph{IEEE Transactions on Pattern Analysis and Machine Intelligence}, vol.~45, no.~5, pp. 6552--6574, 2022.

\bibitem{hoffmann2020real}
J.~E. Hoffmann, H.~G. Tosso, M.~M.~D. Santos, J.~F. Justo, A.~W. Malik, and A.~U. Rahman, ``Real-time adaptive object detection and tracking for autonomous vehicles,'' \emph{IEEE Transactions on Intelligent Vehicles}, vol.~6, no.~3, pp. 450--459, 2020.

\bibitem{liang2021deep}
Y.~Liang, Q.~Wu, Y.~Liu, Y.~Yan, and H.~Wang, ``Deep correlation filter tracking with shepherded instance-aware proposals,'' \emph{IEEE Transactions on Intelligent Transportation Systems}, 2021.

\bibitem{dimp}
G.~Bhat, M.~Danelljan, L.~V. Gool, and R.~Timofte, ``Learning discriminative model prediction for tracking,'' in \emph{Proceedings of the IEEE/CVF international conference on computer vision}, 2019, pp. 6182--6191.

\bibitem{kys}
G.~Bhat, M.~Danelljan, L.~Van~Gool, and R.~Timofte, ``Know your surroundings: Exploiting scene information for object tracking,'' in \emph{European Conference on Computer Vision}.\hskip 1em plus 0.5em minus 0.4em\relax Springer, 2020, pp. 205--221.

\bibitem{atom}
M.~Danelljan, G.~Bhat, F.~S. Khan, and M.~Felsberg, ``Atom: Accurate tracking by overlap maximization,'' in \emph{Proceedings of the IEEE/CVF conference on computer vision and pattern recognition}, 2019, pp. 4660--4669.

\bibitem{danelljan2015learning}
M.~Danelljan, G.~Hager, F.~Shahbaz~Khan, and M.~Felsberg, ``Learning spatially regularized correlation filters for visual tracking,'' in \emph{Proceedings of the IEEE international conference on computer vision}, 2015, pp. 4310--4318.

\bibitem{danelljan2016beyond}
M.~Danelljan, A.~Robinson, F.~Shahbaz~Khan, and M.~Felsberg, ``Beyond correlation filters: Learning continuous convolution operators for visual tracking,'' in \emph{Computer Vision--ECCV 2016: 14th European Conference, Amsterdam, The Netherlands, October 11-14, 2016, Proceedings, Part V 14}.\hskip 1em plus 0.5em minus 0.4em\relax Springer, 2016, pp. 472--488.

\bibitem{visiontransformer2020}
A.~Dosovitskiy, L.~Beyer, A.~Kolesnikov, D.~Weissenborn, X.~Zhai, T.~Unterthiner, M.~Dehghani, M.~Minderer, G.~Heigold, S.~Gelly \emph{et~al.}, ``An image is worth 16x16 words: Transformers for image recognition at scale,'' \emph{arXiv preprint arXiv:2010.11929}, 2020.

\bibitem{wu2021cvt}
H.~Wu, B.~Xiao, N.~Codella, M.~Liu, X.~Dai, L.~Yuan, and L.~Zhang, ``Cvt: Introducing convolutions to vision transformers,'' in \emph{Proceedings of the IEEE/CVF international conference on computer vision}, 2021, pp. 22--31.

\bibitem{siamfc}
L.~Bertinetto, J.~Valmadre, J.~F. Henriques, A.~Vedaldi, and P.~H. Torr, ``Fully-convolutional siamese networks for object tracking,'' in \emph{European Conference on Computer Vision}.\hskip 1em plus 0.5em minus 0.4em\relax Springer, 2016, pp. 850--865.

\bibitem{chen2021transt}
X.~Chen, B.~Yan, J.~Zhu, D.~Wang, X.~Yang, and H.~Lu, ``Transformer tracking,'' in \emph{Proceedings of the IEEE/CVF Conference on Computer Vision and Pattern Recognition}, 2021, pp. 8126--8135.

\bibitem{siamrpn}
B.~Li, J.~Yan, W.~Wu, Z.~Zhu, and X.~Hu, ``High performance visual tracking with siamese region proposal network,'' in \emph{Proceedings of the IEEE/CVF conference on computer vision and pattern recognition}, 2018, pp. 8971--8980.

\bibitem{stark}
B.~Yan, H.~Peng, J.~Fu, D.~Wang, and H.~Lu, ``Learning spatio-temporal transformer for visual tracking,'' in \emph{Proceedings of the IEEE/CVF International Conference on Computer Vision}, 2021, pp. 10\,448--10\,457.

\bibitem{chen2022SimTrack}
B.~Chen, P.~Li, L.~Bai, L.~Qiao, Q.~Shen, B.~Li, W.~Gan, W.~Wu, and W.~Ouyang, ``Backbone is all your need: A simplified architecture for visual object tracking,'' \emph{ECCV}, 2022.

\bibitem{chen2023seqtrack}
X.~Chen, H.~Peng, D.~Wang, H.~Lu, and H.~Hu, ``Seqtrack: Sequence to sequence learning for visual object tracking,'' in \emph{Proceedings of the IEEE/CVF Conference on Computer Vision and Pattern Recognition}, 2023, pp. 14\,572--14\,581.

\bibitem{mixformer}
Y.~Cui, C.~Jiang, L.~Wang, and G.~Wu, ``Mixformer: End-to-end tracking with iterative mixed attention,'' in \emph{Proceedings of the IEEE/CVF Conference on Computer Vision and Pattern Recognition}, 2022, pp. 13\,608--13\,618.

\bibitem{xie2022sbt}
F.~Xie, C.~Wang, G.~Wang, Y.~Cao, W.~Yang, and W.~Zeng, ``Correlation-aware deep tracking,'' in \emph{Proceedings of the IEEE/CVF Conference on Computer Vision and Pattern Recognition}, 2022, pp. 8751--8760.

\bibitem{ye2022OSTrack}
B.~Ye, H.~Chang, B.~Ma, S.~Shan, and X.~Chen, ``Joint feature learning and relation modeling for tracking: A one-stream framework,'' in \emph{European conference on computer vision}.\hskip 1em plus 0.5em minus 0.4em\relax Springer, 2022, pp. 341--357.

\bibitem{chen2023vision}
\BIBentryALTinterwordspacing
Z.~Chen, Y.~Duan, W.~Wang, J.~He, T.~Lu, J.~Dai, and Y.~Qiao, ``Vision transformer adapter for dense predictions,'' in \emph{The Eleventh International Conference on Learning Representations}, 2023. [Online]. Available: \url{https://openreview.net/forum?id=plKu2GByCNW}
\BIBentrySTDinterwordspacing

\bibitem{liu2021swin}
Z.~Liu, Y.~Lin, Y.~Cao, H.~Hu, Y.~Wei, Z.~Zhang, S.~Lin, and B.~Guo, ``Swin transformer: Hierarchical vision transformer using shifted windows,'' in \emph{Proceedings of the IEEE/CVF international conference on computer vision}, 2021, pp. 10\,012--10\,022.

\bibitem{wang2021pyramid}
W.~Wang, E.~Xie, X.~Li, D.-P. Fan, K.~Song, D.~Liang, T.~Lu, P.~Luo, and L.~Shao, ``Pyramid vision transformer: A versatile backbone for dense prediction without convolutions,'' in \emph{Proceedings of the IEEE/CVF international conference on computer vision}, 2021, pp. 568--578.

\bibitem{xie2021segformer}
E.~Xie, W.~Wang, Z.~Yu, A.~Anandkumar, J.~M. Alvarez, and P.~Luo, ``Segformer: Simple and efficient design for semantic segmentation with transformers,'' \emph{Advances in Neural Information Processing Systems}, vol.~34, pp. 12\,077--12\,090, 2021.

\bibitem{Fan_2020_IJCV_Lasot_ext}
H.~Fan, H.~Bai, L.~Lin, F.~Yang, P.~Chu, G.~Deng, S.~Yu, M.~Huang, J.~Liu, Y.~Xu \emph{et~al.}, ``Lasot: A high-quality large-scale single object tracking benchmark,'' \emph{International Journal of Computer Vision (IJCV)}, vol. 129, no.~2, pp. 439--461, 2021.

\bibitem{lasot}
H.~Fan, L.~Lin, F.~Yang, P.~Chu, G.~Deng, S.~Yu, H.~Bai, Y.~Xu, C.~Liao, and H.~Ling, ``Lasot: A high-quality benchmark for large-scale single object tracking,'' in \emph{Proceedings of the IEEE/CVF conference on computer vision and pattern recognition}, 2019, pp. 5374--5383.

\bibitem{survey_fahad}
S.~Javed, M.~Danelljan, F.~S. Khan, M.~H. Khan, M.~Felsberg, and J.~Matas, ``Visual object tracking with discriminative filters and siamese networks: a survey and outlook,'' \emph{IEEE Transactions on Pattern Analysis and Machine Intelligence}, vol.~45, no.~5, pp. 6552--6574, 2022.

\bibitem{survey2}
S.~M. Marvasti-Zadeh, L.~Cheng, H.~Ghanei-Yakhdan, and S.~Kasaei, ``Deep learning for visual tracking: A comprehensive survey,'' \emph{IEEE Transactions on Intelligent Transportation Systems}, vol.~23, no.~5, pp. 3943--3968, 2022.

\bibitem{siamban}
Z.~Chen, B.~Zhong, G.~Li, S.~Zhang, and R.~Ji, ``Siamese box adaptive network for visual tracking,'' in \emph{Proceedings of the IEEE/CVF conference on computer vision and pattern recognition}, 2020, pp. 6668--6677.

\bibitem{li2019siamrpnpp}
B.~Li, W.~Wu, Q.~Wang, F.~Zhang, J.~Xing, and J.~Yan, ``Siamrpn++: Evolution of siamese visual tracking with very deep networks,'' in \emph{Proceedings of the IEEE/CVF Conference on Computer Vision and Pattern Recognition}, 2019, pp. 4282--4291.

\bibitem{carion2020detr}
N.~Carion, F.~Massa, G.~Synnaeve, N.~Usunier, A.~Kirillov, and S.~Zagoruyko, ``End-to-end object detection with transformers,'' in \emph{European conference on computer vision}.\hskip 1em plus 0.5em minus 0.4em\relax Springer, 2020, pp. 213--229.

\bibitem{vaswani2017attention}
A.~Vaswani, N.~Shazeer, N.~Parmar, J.~Uszkoreit, L.~Jones, A.~N. Gomez, {\L}.~Kaiser, and I.~Polosukhin, ``Attention is all you need,'' \emph{Advances in neural information processing systems}, vol.~30, 2017.

\bibitem{gao2022aiatrack}
S.~Gao, C.~Zhou, C.~Ma, X.~Wang, and J.~Yuan, ``Aiatrack: Attention in attention for transformer visual tracking,'' \emph{arXiv preprint arXiv:2207.09603}, 2022.

\bibitem{trdimp}
N.~Wang, W.~Zhou, J.~Wang, and H.~Li, ``Transformer meets tracker: Exploiting temporal context for robust visual tracking,'' in \emph{Proceedings of the IEEE/CVF conference on computer vision and pattern recognition}, 2021, pp. 1571--1580.

\bibitem{prdimp}
M.~Danelljan, L.~V. Gool, and R.~Timofte, ``Probabilistic regression for visual tracking,'' in \emph{Proceedings of the IEEE/CVF conference on computer vision and pattern recognition}, 2020, pp. 7183--7192.

\bibitem{keeptrack}
C.~Mayer, M.~Danelljan, D.~P. Paudel, and L.~Van~Gool, ``Learning target candidate association to keep track of what not to track,'' in \emph{Proceedings of the IEEE/CVF International Conference on Computer Vision}, 2021, pp. 13\,444--13\,454.

\bibitem{tomp}
C.~Mayer, M.~Danelljan, G.~Bhat, M.~Paul, D.~P. Paudel, F.~Yu, and L.~Van~Gool, ``Transforming model prediction for tracking,'' in \emph{Proceedings of the IEEE/CVF Conference on Computer Vision and Pattern Recognition}, 2022, pp. 8731--8740.

\bibitem{yan2021lighttrack}
B.~Yan, H.~Peng, K.~Wu, D.~Wang, J.~Fu, and H.~Lu, ``Lighttrack: Finding lightweight neural networks for object tracking via one-shot architecture search,'' in \emph{Proceedings of the IEEE/CVF Conference on Computer Vision and Pattern Recognition}, 2021, pp. 15\,180--15\,189.

\bibitem{borsuk2022fear}
V.~Borsuk, R.~Vei, O.~Kupyn, T.~Martyniuk, I.~Krashenyi, and J.~Matas, ``Fear: Fast, efficient, accurate and robust visual tracker,'' in \emph{European Conference on Computer Vision}.\hskip 1em plus 0.5em minus 0.4em\relax Springer, 2022, pp. 644--663.

\bibitem{kang2023hit}
B.~Kang, X.~Chen, D.~Wang, H.~Peng, and H.~Lu, ``Exploring lightweight hierarchical vision transformers for efficient visual tracking,'' in \emph{Proceedings of the IEEE/CVF International Conference on Computer Vision}, 2023, pp. 9612--9621.

\bibitem{ettrack}
P.~Blatter, M.~Kanakis, M.~Danelljan, and L.~Van~Gool, ``Efficient visual tracking with exemplar transformers,'' in \emph{Proceedings of the IEEE/CVF Winter conference on applications of computer vision}, 2023, pp. 1571--1581.

\bibitem{gao2023GRM}
S.~Gao, C.~Zhou, and J.~Zhang, ``Generalized relation modeling for transformer tracking,'' in \emph{Proceedings of the IEEE/CVF Conference on Computer Vision and Pattern Recognition}, 2023, pp. 18\,686--18\,695.

\bibitem{henriques2014high}
J.~F. Henriques, R.~Caseiro, P.~Martins, and J.~Batista, ``High-speed tracking with kernelized correlation filters,'' \emph{IEEE Transactions on pattern analysis and machine intelligence}, vol.~37, no.~3, pp. 583--596, 2014.

\bibitem{ltmu}
K.~Dai, Y.~Zhang, D.~Wang, J.~Li, H.~Lu, and X.~Yang, ``High-performance long-term tracking with meta-updater,'' in \emph{Proceedings of the IEEE/CVF Conference on Computer Vision and Pattern Recognition}, 2020, pp. 6298--6307.

\bibitem{zhang2019learning}
L.~Zhang, A.~Gonzalez-Garcia, J.~V.~D. Weijer, M.~Danelljan, and F.~S. Khan, ``Learning the model update for siamese trackers,'' in \emph{Proceedings of the IEEE/CVF international conference on computer vision}, 2019, pp. 4010--4019.

\bibitem{jiang2018iounet}
B.~Jiang, R.~Luo, J.~Mao, T.~Xiao, and Y.~Jiang, ``Acquisition of localization confidence for accurate object detection,'' in \emph{Proceedings of the European conference on computer vision (ECCV)}, 2018, pp. 784--799.

\bibitem{dai2019visual}
K.~Dai, D.~Wang, H.~Lu, C.~Sun, and J.~Li, ``Visual tracking via adaptive spatially-regularized correlation filters,'' in \emph{Proceedings of the IEEE/CVF Conference on Computer Vision and Pattern Recognition}, 2019, pp. 4670--4679.

\bibitem{zheng2020learning}
L.~Zheng, M.~Tang, Y.~Chen, J.~Wang, and H.~Lu, ``Learning feature embeddings for discriminant model based tracking,'' in \emph{Computer Vision--ECCV 2020: 16th European Conference, Glasgow, UK, August 23--28, 2020, Proceedings, Part XV 16}.\hskip 1em plus 0.5em minus 0.4em\relax Springer, 2020, pp. 759--775.

\bibitem{nt2019revisiting}
H.~Nt and T.~Maehara, ``Revisiting graph neural networks: All we have is low-pass filters,'' \emph{arXiv preprint arXiv:1905.09550}, 2019.

\bibitem{cai2020note}
C.~Cai and Y.~Wang, ``A note on over-smoothing for graph neural networks,'' \emph{arXiv preprint arXiv:2006.13318}, 2020.

\bibitem{tang2023learning}
C.~Tang, X.~Wang, Y.~Bai, Z.~Wu, J.~Zhang, and Y.~Huang, ``Learning spatial-frequency transformer for visual object tracking,'' \emph{IEEE Transactions on Circuits and Systems for Video Technology}, 2023.

\bibitem{Yan_2021_CVPR_AlphaRefine}
B.~Yan, X.~Zhang, D.~Wang, H.~Lu, and X.~Yang, ``Alpha-refine: Boosting tracking performance by precise bounding box estimation,'' in \emph{Proceedings of the IEEE/CVF Conference on Computer Vision and Pattern Recognition}, 2021, pp. 5289--5298.

\bibitem{siamr-cnn}
P.~Voigtlaender, J.~Luiten, P.~H. Torr, and B.~Leibe, ``Siam r-cnn: Visual tracking by re-detection,'' in \emph{Proceedings of the IEEE/CVF conference on computer vision and pattern recognition}, 2020, pp. 6578--6588.

\bibitem{Danelljan_2019_github_pytracking}
M.~Danelljan and G.~Bhat, ``{PyTracking: Visual tracking library based on PyTorch.}'' \url{https://github.com/visionml/pytracking}, 2019, accessed: 1/05/2021.

\bibitem{coco}
T.-Y. Lin, M.~Maire, S.~Belongie, J.~Hays, P.~Perona, D.~Ramanan, P.~Doll{\'a}r, and C.~L. Zitnick, ``Microsoft coco: Common objects in context,'' in \emph{European Conference on Computer Vision}.\hskip 1em plus 0.5em minus 0.4em\relax Springer, 2014, pp. 740--755.

\bibitem{got10k}
L.~Huang, X.~Zhao, and K.~Huang, ``Got-10k: A large high-diversity benchmark for generic object tracking in the wild,'' \emph{IEEE Transactions on Pattern Analysis and Machine Intelligence}, 2019.

\bibitem{trackingnet}
M.~Muller, A.~Bibi, S.~Giancola, S.~Alsubaihi, and B.~Ghanem, ``Trackingnet: A large-scale dataset and benchmark for object tracking in the wild,'' in \emph{European Conference on Computer Vision}, 2018, pp. 300--317.

\bibitem{he2022mae}
K.~He, X.~Chen, S.~Xie, Y.~Li, P.~Doll{\'a}r, and R.~Girshick, ``Masked autoencoders are scalable vision learners,'' in \emph{Proceedings of the IEEE/CVF conference on computer vision and pattern recognition}, 2022, pp. 16\,000--16\,009.

\bibitem{adamw}
I.~Loshchilov and F.~Hutter, ``Decoupled weight decay regularization,'' \emph{arXiv preprint arXiv:1711.05101}, 2017.

\bibitem{Rezatofighi_2019_CVPR_GIOU}
H.~Rezatofighi, N.~Tsoi, J.~Gwak, A.~Sadeghian, I.~Reid, and S.~Savarese, ``Generalized intersection over union: A metric and a loss for bounding box regression,'' in \emph{Proceedings of the IEEE/CVF conference on computer vision and pattern recognition}, 2019, pp. 658--666.

\bibitem{swintrack}
L.~Lin, H.~Fan, Y.~Xu, and H.~Ling, ``Swintrack: A simple and strong baseline for transformer tracking,'' \emph{arXiv preprint arXiv:2112.00995}, 2021.

\bibitem{automatch}
Z.~Zhang, Y.~Liu, X.~Wang, B.~Li, and W.~Hu, ``Learn to match: Automatic matching network design for visual tracking,'' in \emph{Proceedings of the IEEE international conference on computer vision}, 2021, pp. 13\,339--13\,348.

\bibitem{noman2022avist}
M.~Noman, W.~A. Ghallabi, D.~Najiha, C.~Mayer, A.~Dudhane, M.~Danelljan, H.~Cholakkal, S.~Khan, L.~Van~Gool, and F.~S. Khan, ``Avist: A benchmark for visual object tracking in adverse visibility,'' \emph{arXiv preprint arXiv:2208.06888}, 2022.

\bibitem{paul2022rts}
M.~Paul, M.~Danelljan, C.~Mayer, and L.~Van~Gool, ``Robust visual tracking by segmentation,'' \emph{arXiv preprint arXiv:2203.11191}, 2022.

\bibitem{zhang2020ocean}
Z.~Zhang, H.~Peng, J.~Fu, B.~Li, and W.~Hu, ``Ocean: Object-aware anchor-free tracking,'' in \emph{Computer Vision--ECCV 2020: 16th European Conference, Glasgow, UK, August 23--28, 2020, Proceedings, Part XXI 16}.\hskip 1em plus 0.5em minus 0.4em\relax Springer, 2020, pp. 771--787.

\bibitem{Kristan_2020_ECCVW_VOT2020}
M.~Kristan, A.~Leonardis, J.~Matas, M.~Felsberg, R.~Pflugfelder, J.-K. K{\"a}m{\"a}r{\"a}inen, M.~Danelljan, L.~{\v{C}}. Zajc, A.~Luke{\v{z}}i{\v{c}}, O.~Drbohlav \emph{et~al.}, ``The eighth visual object tracking vot2020 challenge results,'' in \emph{Computer Vision--ECCV 2020 Workshops: Glasgow, UK, August 23--28, 2020, Proceedings, Part V 16}.\hskip 1em plus 0.5em minus 0.4em\relax Springer, 2020, pp. 547--601.

\bibitem{cswintt}
Z.~Song, J.~Yu, Y.-P.~P. Chen, and W.~Yang, ``Transformer tracking with cyclic shifting window attention,'' in \emph{Proceedings of the IEEE/CVF Conference on Computer Vision and Pattern Recognition}, 2022, pp. 8791--8800.

\bibitem{Mueller_2016_ECCV_UAV123}
M.~Mueller, N.~Smith, and B.~Ghanem, ``A benchmark and simulator for uav tracking,'' in \emph{Computer Vision--ECCV 2016: 14th European Conference, Amsterdam, The Netherlands, October 11--14, 2016, Proceedings, Part I 14}.\hskip 1em plus 0.5em minus 0.4em\relax Springer, 2016, pp. 445--461.

\bibitem{tnl2k}
X.~Wang, X.~Shu, Z.~Zhang, B.~Jiang, Y.~Wang, Y.~Tian, and F.~Wu, ``Towards more flexible and accurate object tracking with natural language: Algorithms and benchmark,'' in \emph{Proceedings of the IEEE/CVF conference on computer vision and pattern recognition}, 2021, pp. 13\,763--13\,773.

\bibitem{fu2021stmtrack}
Z.~Fu, Q.~Liu, Z.~Fu, and Y.~Wang, ``Stmtrack: Template-free visual tracking with space-time memory networks,'' in \emph{Proceedings of the IEEE/CVF conference on computer vision and pattern recognition}, 2021, pp. 13\,774--13\,783.

\bibitem{Matej_2018_ECCVW_VOT2018}
M.~Kristan, A.~Leonardis, J.~Matas, M.~Felsberg, R.~Pflugfelder, L.~ˇCehovin~Zajc, T.~Vojir, G.~Bhat, A.~Lukezic, A.~Eldesokey \emph{et~al.}, ``The sixth visual object tracking vot2018 challenge results,'' in \emph{Proceedings of the European conference on computer vision (ECCV) workshops}, 2018, pp. 0--0.

\bibitem{Kristan_2019_ICCVW_VOT2019}
M.~Kristan, J.~Matas, A.~Leonardis, M.~Felsberg, R.~Pflugfelder, J.-K. Kamarainen, L.~ˇCehovin~Zajc, O.~Drbohlav, A.~Lukezic, A.~Berg \emph{et~al.}, ``The seventh visual object tracking vot2019 challenge results,'' in \emph{Proceedings of the IEEE/CVF international conference on computer vision workshops}, 2019, pp. 0--0.

\bibitem{ke2023samhq}
L.~Ke, M.~Ye, M.~Danelljan, Y.~Liu, Y.-W. Tai, C.-K. Tang, and F.~Yu, ``Segment anything in high quality,'' \emph{arXiv preprint arXiv:2306.01567}, 2023.

\end{thebibliography}
